\documentclass[lettersize,journal]{IEEEtran}
\usepackage{amsmath,amsfonts}
\usepackage{algorithmic}
\usepackage{algorithm}
\usepackage{array}
\usepackage[caption=false,font=normalsize,labelfont=sf,textfont=sf]{subfig}
\usepackage{textcomp}
\usepackage{stfloats}
\usepackage{url}
\usepackage{verbatim}
\usepackage{graphicx}
\usepackage{multirow}
\usepackage[table,xcdraw]{xcolor}

\begin{document}

\title{Cross-domain Random Pre-training with Prototypes for Reinforcement Learning}

\author{Xin Liu,~\IEEEmembership{Student Member,~IEEE,} Yaran Chen,~\IEEEmembership{Member,~IEEE,} Haoran Li,~\IEEEmembership{Member,~IEEE,} 
Boyu Li,~\IEEEmembership{Student Member,~IEEE,} and Dongbin Zhao,~\IEEEmembership{Fellow,~IEEE}
        % <-this % stops a space

\thanks{This work has been submitted to the IEEE for possible publication. Copyright may be transferred without notice, after which this version may no longer be accessible. Implementation of this work will be arranged and released after publication.}

\thanks{X. Liu, Y. Chen, H. Li, B. Li, and D. Zhao are with the State Key Laboratory
of Multimodal Artificial Intelligence Systems, Institute of Automation, Chinese Academy of Sciences, Beijing 100190, and also with the School of
Artificial Intelligence, University of Chinese Academy of Sciences, Beijing
100049, China. (email: liuxin2021@ia.ac.cn, chenyaran2013@ia.ac.cn, lihaoran2015@ia.ac.cn, liboyu2021@ia.ac.cn, dongbin.zhao@ia.ac.cn)} 

\thanks{B. Li is also with the Beijing Academy of Artificial Intelligence, Beijing, China.}% <-this % stops a space    , Beijing, 100084, China.

%\thanks{Yaran Chen and Dongbin Zhao are both corresponding authors.}

% <-this % stops a space
}

\markboth{Journal of \LaTeX\ Class Files, IEEE Transactions on Systems, Man and Cybernetics: Systems}%
{Shell \MakeLowercase{\textit{et al.}}: A Sample Article Using IEEEtran.cls for IEEE Journals}



\maketitle

\begin{abstract}
Unsupervised cross-domain Reinforcement Learning (RL) pre-training shows great potential for challenging continuous visual control but poses a big challenge. In this paper, we propose \textbf{C}ross-domain \textbf{R}andom \textbf{P}re-\textbf{T}raining with \textbf{pro}totypes (CRPTpro), a novel, efficient, and effective self-supervised cross-domain RL pre-training framework. CRPTpro decouples data sampling from encoder pre-training, proposing decoupled random collection to easily and quickly generate a qualified cross-domain pre-training dataset. Moreover, a novel prototypical self-supervised algorithm is proposed to pre-train an effective visual encoder that is generic across different domains. Without finetuning, the cross-domain encoder can be implemented for challenging downstream tasks defined in different domains, either seen or unseen. Compared with recent advanced methods, CRPTpro achieves better performance on downstream policy learning without extra training on exploration agents for data collection, greatly reducing the burden of pre-training. We conduct extensive experiments across eight challenging continuous visual-control domains, including balance control, robot locomotion, and manipulation. CRPTpro significantly outperforms the next best Proto-RL(C) on 11/12 cross-domain downstream tasks with only 54.5\% wall-clock pre-training time, exhibiting state-of-the-art pre-training performance with greatly improved pre-training efficiency. 
\end{abstract}


\section{Introduction}
\IEEEPARstart{R}{epresentation} learning is crucial for Deep Reinforcement Learning (DRL), especially image-based RL. Traditional task-specific RL algorithms~\cite{r14,dreamer,r29,r31} rely on the reward function to learn feature representation and downstream policy simultaneously \cite{smc-1,r15,smc-pre-training,r44}, succeeding in many fields~\cite{smc-2,r45,r47,smc-3}. However, it is sample inefficient when faced with high-dimensional input like images, especially complex visual input in challenging continuous visual motor control. %The reward signal does not directly judge the representation ability of the visual encoder. When input is complex or reward is sparse, it is difficult to train an ideal visual encoder relying on only reward function, let alone downstream policy learning. 
Besides, the task-specified encoder learned with lots of effort cannot generalize to novel tasks. %It is narrowed to a single task. %Even for similar tasks defined in the same domain, it is hard for the task-specified encoder to generalize well.
For problems mentioned above, self-supervised task-agnostic pre-training, the combination of Self-Supervised Learning (SSL) and unsupervised visual pre-training, is proposed. By designing auxiliary tasks, SSL improves the perception ability of the visual encoder in a targeted manner, thus learning better representation for downstream policy. Employing SSL to achieve pre-training over a task-agnostic dataset makes the encoder generic across different downstream tasks. Recent works~\cite{r9,r12} have proven it possible to pre-train a powerful single-domain encoder which enables efficient downstream RL on different challenging visual-control tasks defined in the same domain.
%Self-Supervised Learning (SSL), which demonstrates its strong representation capabilities in the fields of Computer Vision (CV) and Natural Language Processing (NLP) is introduced into DRL.  Initially, the self-supervised DRL algorithms are end-to-end, which introduce the auxiliary loss additionally to the RL loss. The encoders are updated by both auxiliary loss and RL loss, which narrows their utility to a single task. Then, inspired by the recent success of task-agnostic visual learning in CV, some works try to decouple representation learning from RL, proposing encoder pre-training for DRL. by cutting off the backpropagation from RL loss to visual encoder. This decoupling is not complete because the data for the auxiliary task is collected during the policy learning, which means representation and policy are still learned simultaneously. For the second cause mentioned above, more and more recent works focus on self-supervised pre-training for RL. They pre-train a visual encoder for downstream policy learning over a task-agnostic dataset, achieving complete decoupling from representation learning to policy learning. 
%For the second problem, task-agnostic pre-training is an ideal solution. The encoder is pre-trained on task-agnostic dataset, which means it can generalize in different tasks In addition to the improved sample efficiency, the task-agnostic pre-training brings another benefit: The pre-trained encoder can generalize across different tasks defined in the same domain.
\begin{figure}[t]
    \centering
    \includegraphics[scale=0.42]{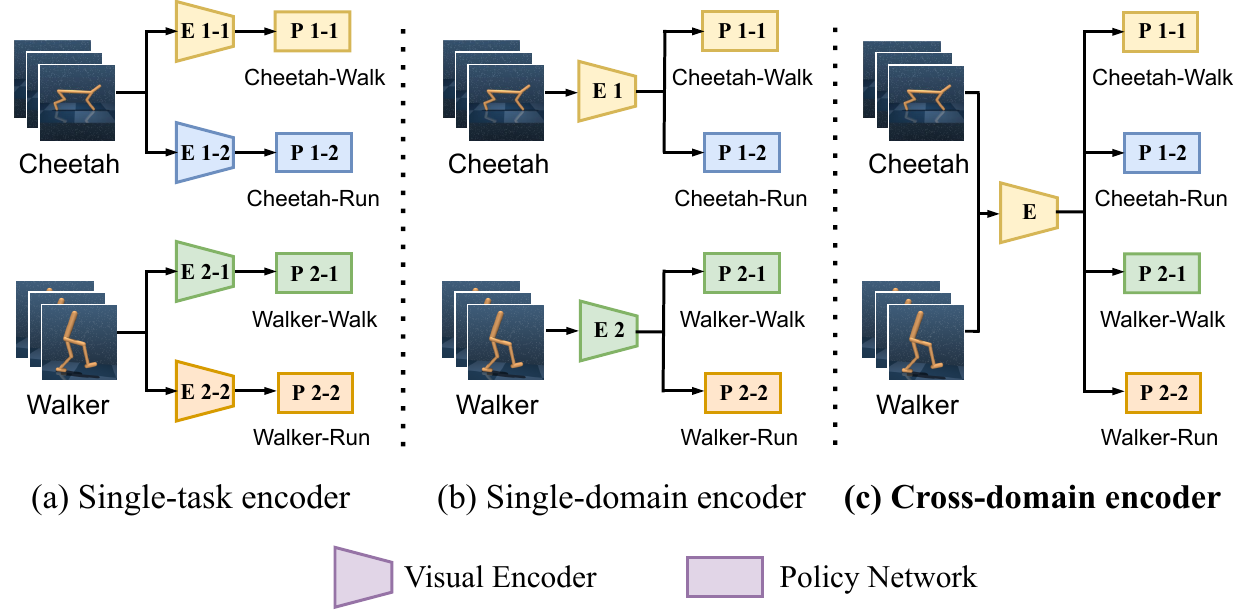}
    \vspace{-5mm}
    \caption{Difference between three kinds of visual encoders in image-based RL on DMControl. (a) The single-task encoder. It is dedicated to only one task. (b) The single-domain encoder. It generalizes across tasks in a single domain, i.e., tasks from the same domain can share the same encoder. \textbf{(c) The cross-domain encoder.} It generalizes across both tasks and domains, i.e., tasks from different domains can share the same encoder. CRPTpro pre-trains a powerful cross-domain encoder enabling state-of-the-art downstream policy learning across multiple domains in challenging continuous visual control.}
    \label{difference}
\end{figure}

\IEEEpubidadjcol

Now that the single-domain encoder has been implemented, another question naturally arises: Can we pre-train a cross-domain encoder that is generic across both tasks and domains? This means that different tasks defined in different domains can share the same encoder. To avoid confusion, we use DeepMind Control suite (DMControl)~\cite{r13} as an example to illustrate the differences between these encoders, as shown in Fig. \ref{difference}. There are many benefits of pre-training a cross-domain encoder. Intuitively, its versatility can greatly reduce the training burden when faced with multiple domains. In the long run, a generic encoder is necessary for many promising sub-fields of DRL, such as multi-tasks RL~\cite{r46}, meta-RL~\cite{r38}, transfer RL~\cite{r41}, and the long-term goal of RL: generalist agent~\cite{r39}. 

Currently, unsupervised active pre-training methods~\cite{r9,r12} are state-of-the-art% on visual-control tasks
, enabling effective encoder pre-training for multiple tasks defined in a single domain. As a branch of unsupervised RL ~\cite{diayn,exploration2,cic-nips,becl}, these approaches train extra exploration agents via unsupervised intrinsic reward to collect task-agnostic data for encoder learning. The exploration policy is built on the visual encoder, and they are updated simultaneously during the exploration agent training (i.e., pre-training). However, two problems emerge when transplanting these methods into cross-domain pre-training. First, the pre-trained encoder degenerates, leading to poor downstream policy learning. In cross-domain settings, the chicken-and-egg problem between exploration agent training and visual encoder pre-training is much more severe than that in single-domain pre-training. It's difficult for current active methods to train multiple qualified exploration agents on a substandard encoder that is under cross-domain pre-training, and vice versa. Second, the pre-training efficiency is insufficient. As mentioned above, current advanced active pre-training methods employ extra RL to train exploration agents for data collection. This leads to a severe pre-training burden, which multiplies with the number of domains.

In this work, we address the above problems through \textbf{CRPTpro}, a novel, efficient, and effective self-supervised cross-domain RL pre-training framework. % for visual-control tasks. 
%CRPTpro can efficiently pre-train an effective visual encoder which is able to conduct efficient downstream RL on ch . 
Instead of training extra exploration agents for data collection like recent advanced unsupervised methods, CRPTpro completely decouples data sampling from encoder pre-training, proposing decoupled random collection. It employs an off-the-shelf random policy to achieve steady exploration across multiple domains, easily and quickly producing a qualified cross-domain pre-training dataset to improve both the performance and efficiency of encoder pre-training. 
Moreover, a novel self-supervised algorithm named efficient prototypical learning is proposed to achieve advanced encoder pre-training over the sampled dataset. It projects observation encodings onto some trainable prototypes that serve as the cluster centers and compares them with clustering assignment targets, while facilitating the diffusion and coverage of prototypes.
%Moreover, a novel self-supervised algorithm named efficient prototypical learning is proposed to further improve the pre-training performance. It improves original prototypical representation learning~\cite{r24,r12,r26,r48} by a novel intrinsic loss that facilitates the diffusion of prototypes. 
After pre-training, the cross-domain encoder obtained by CRPTpro is frozen and achieves efficient downstream policy learning via RL on challenging visual-control tasks from different domains. Besides, the pre-trained encoder can generalize well to unseen domains directly or through only a few-shot finetuning. %In addition, the cross-domain encoder could be finetuned on a specified domain, which
%makes the encoder dedicated to only one domain, 
%reduces its versatility but further improves its performance, especially in unseen domains. 
We conduct extensive experiments on eight different, representative and challenging continuous visual-control environments, including classical balance control like pendulum, multi-joint robot locomotion, robotic manipulation and so on. % from DMControl ~\cite{r13}, a representative benchmark for continuous visual-control tasks which are most challenging for unsupervised RL. 
Results demonstrate that CRPTpro outperforms all cross-domain pre-training baselines significantly, enabling state-of-the-art cross-domain downstream policy learning. % that is even competitive with the best image-based RL algorithm. 
For pre-training efficiency, CRPTpro also exceeds the most advanced approach by a large margin. In addition, we demonstrate and analyze the effectiveness of our decoupled random collection and efficient prototypical learning in different pre-training settings.

The contributions of our paper can be summarized as:
\begin{itemize}
    \item We propose CRPTpro, a novel, efficient, and effective self-supervised cross-domain RL pre-training framework. The cross-domain encoder obtained by CRPTpro enables efficient downstream RL on different challenging visual-control tasks defined in different domains. In addition, it can generalize well to unseen domains directly or through a few-shot finetuning.
    \item Unlike recent advanced methods, CRPTpro decouples data sampling from encoder pre-training, proposing decoupled random collection to easily and quickly generate a qualified cross-domain pre-training dataset, which improves both pre-training efficiency and downstream policy performance. In addition, CRPTpro proposes efficient prototypical learning, a novel prototypical self-supervised algorithm to further improve the pre-training.
    \item Extensive experiments demonstrate that CRPTpro significantly outperforms all cross-domain baselines on downstream policy learning. It improves current state-of-the-art method by a large margin, with greatly improved pre-training efficiency. In addition, we provide detailed analysis about the proposed decoupled random collection and efficient prototypical learning.
\end{itemize}

\section{Related Work}
\subsection{Self-supervised Learning in RL}

End-to-end RL shows sample inefficiency when faced with high-dimensional observations like images. One solution is to employ auxiliary objectives to do SSL, for example, predicting the designed properties of the environments~\cite{r18,r19}. % or using auto-encoder ~\cite{r36,r17}. %Following these, SAC-AE (36) and SLAC (17) use auto-encoder to improve representation learning in RL. 
Following these, CURL~\cite{r20} introduced a contrastive auxiliary task into end-to-end RL to improve sample efficiency. SPR~\cite{r21} combined data augmentation with an auxiliary SSL objective. 
%ATC proposed a temporal contrastive self-supervised learning algorithm for both end-to-end RL and encoder pretraining. Its success to decouple representation learning from task reward inspired lots of excellent works about encoder pretraining, which we would introduce in the next part. 
CltrFormer~\cite{r41} employed transformer~\cite{r25} in visual-control tasks, using contrastive learning to get transferable representation between different domains on DMControl~\cite{r13}. Recently, a novel SSL approach: prototypical representation learning~\cite{r24} based on the Sinkhorn-Knopp algorithm~\cite{r23} was introduced into RL by Proto-RL~\cite{r12}. Their success motivated DreamerPro~\cite{r26}, ProtoCAD~\cite{r48} and our CRPTpro. Different from previous algorithms, our efficient prototypical learning compares the observations projected onto some prototypes (serving as cluster centers) with their clustering assignment targets, simultaneously facilitating the diffusion of the prototypes. In addition, CRPTpro performs prototypical representation learning on a static dataset as SwAV~\cite{r24} does.

\subsection{Encoder Pre-training for Downstream RL}

Inspired by the success that SSL can pre-train a strong feature extractor without labels in the field of CV~\cite{r1,r3} and NLP~\cite{r4,r5}, ATC~\cite{r6} tried to decouple the representation learning from downstream policy learning and first achieved considerable results. %They proposed a time-contrastive SSL algorithm and pre-train a general encoder frozen for different tasks from different domains in DMC(13). 
%ATC could strengthen both end-to-end RL and encoder pre-training but required expert datasets. 
MVP~\cite{r10} utilized the offline dataset from the internet, pre-training a visual encoder %using MAE ~\cite{r11} 
for different motor tasks. SGI~\cite{r7} employed finetuning on the pre-trained encoder, achieving efficient policy learning on Atari 100k benchmark~\cite{r8}. Inspired by unsupervised RL ~\cite{diayn,urlb,exploration2}, APT~\cite{r9} pre-trained an extra exploration agent via an unsupervised intrinsic reward for data collection, along with visual representation learning by 
SimCLR~\cite{r3}. 
%It could handle both single-domain and cross-domain pre-training but required 5M pretraining steps of huge inefficiency. 
Subsequently, Proto-RL~\cite{r12} used prototypical representation learning~\cite{r24} to simultaneously achieve encoder pre-training and strengthen APT unsupervised reward, setting state-of-the-art downstream policy performance on DMControl. %Similar to APT and Proto-RL, lots of recent pre-training methods are devoted to training a specified agent for data collection, which is time-consuming.
Different from these active pre-training methods~\cite{r9,r12,becl}, CRPTpro gives up unsupervised RL but decouples data collection from encoder pre-training by an off-the-shelf exploration policy, achieving state-of-the-art cross-domain pre-training performance with dramatically improved pre-training efficiency.

\begin{figure*}
    \centering
    \includegraphics[scale=0.42 ]{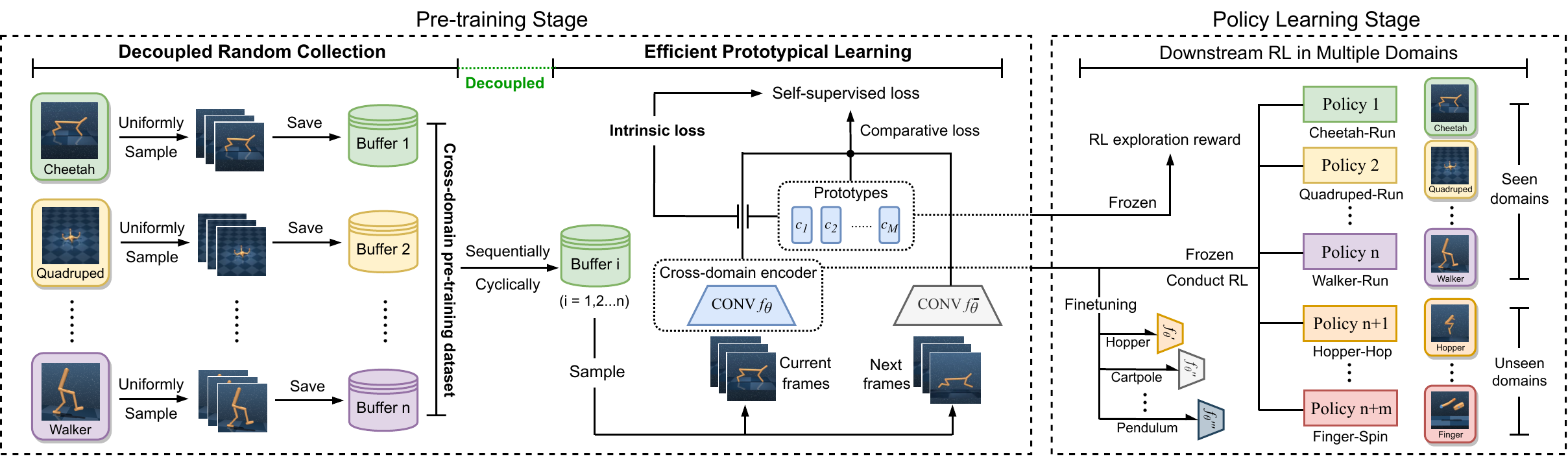}
    \caption{The schematic diagram of CRPTpro. In the pre-training, CRPTpro decouples data sampling from encoder pre-training, employing decoupled random collection to easily and quickly produce a qualified cross-domain pre-training dataset for SSL. Next, CRPTpro employs a novel self-supervised algorithm (efficient prototypical learning) over different data buffers sequentially and cyclically, to pre-train an effective cross-domain encoder and some prototypes. After the pre-training, the encoder and prototypes are frozen and used to perform efficient downstream RL on sets of challenging continuous visual-control tasks from different domains either seen or unseen. Finetuning in a single domain is optional, leading to better single-domain downstream policy learning but reducing the cross-domain versatility of the encoder to some extent.}
    \label{pipeline}
    %\vspace{-5mm}
\end{figure*}

\section{Methodology}
\textit{Problem definition:} In most circumstances, a visual-control task could be formulated as an infinite-horizon Partially Observable Markov Decision Process (POMDP)~\cite{r34,r33}, denoted by $\mathcal{M}^p=(\mathcal{O},\mathcal{A},\mathcal{P},\mathcal{R},\gamma,d_0)$, where $\mathcal{O}$ is the high-dimensional observation space, i.e. pixels, $\mathcal{A}$ is the action space, $\mathcal{P}$ is the distribution of next observation given the history and current action, $R$ is the reward function, $\gamma$ is the discount factor, and $d_0$ is the distribution of the initial observation. By stacking three consecutive previous observations into a state, this POMDP is converted into an Markov Decision Process (MDP) $\mathcal{M}= (\mathcal{X},\mathcal{A},\mathcal{P},\mathcal{R},\gamma,d_0)$, where the next state only depends on the current state, unrelated to the history. RL can be performed on the MDP to obtain a downstream policy. 

The proposed CRPTpro is shown in Fig. \ref{pipeline}. During pre-training, we propose decoupled random collection to efficiently produce an effective cross-domain pre-training dataset, as shown in Section III.A. Then, we describe how the efficient prototypical learning is implemented and how to pre-train the cross-domain encoder and prototypes, which is detailed in Section III.B. After pre-training, the frozen cross-domain encoder and prototypes enable efficient downstream RL on different challenging visual-control tasks from different domains, which we describe in Section III.C.

\subsection{Decoupled Random Collection}
The dataset for pre-training should be diverse to cover the observation space as much as possible. As a branch of unsupervised RL, active pre-training methods~\cite{r9,r12} design an intrinsic reward related to exploration, based on which they train extra agents to explore the observation space along with learning visual encoders. These methods are able to explore and collect far-reaching states in hard-exploration domains and achieve state-of-the-art performance in single-domain visual-control pre-training. However, due to the requirements of extra agent training, these methods suffer a severe pre-training burden that even exceeds that of downstream RL many times. Especially when facing multiple domains, this extra burden is dramatically increased. Meanwhile, these approaches exhibit huge performance drops when transplanted into cross-domain pre-training due to a chicken-and-egg problem: pre-training an effective encoder requires effective exploration agents for qualified data collection, while effective exploration agents rely on an effective encoder. This problem is severely amplified in cross-domain pre-training because multiple exploration policies are required for multiple domains. It's hard for current active methods to train multiple qualified exploration strategies from scratch on a substandard visual encoder, and it's also hard to achieve ideal pre-training on unqualified data sampled by unqualified exploration policies.

Due to the above problems, CRPTpro gives up training extra unsupervised exploration agents but decouples data sampling from encoder pre-training, proposing decoupled random collection for cross-domain pre-training. It employs an off-the-shelf random policy across multiple domains for pre-training dataset collection, which dramatically reduces the pre-training burden. Specifically, it chooses a simple uniform distribution to sample actions for environment interaction and data (image observations) collection. Data from different domains is saved into different data buffers, forming the cross-domain pre-training dataset, which is collected once and used permanently. The capacity of each buffer is $b$. This cross-domain pre-training dataset is then used for SSL. 

In terms of the ability to explore far-reaching states, the uniform distribution policy is not as effective as active exploration. However, for visual-control pre-training across multiple domains, our decoupled random collection performs better than unsupervised active pre-training because it provides qualified training data and a more stable training process. Firstly, the above chicken-and-egg conflict makes it hard to train qualified exploration policies for active exploration, while our off-the-shelf uniform distribution policy is not affected. Secondly, the difficulty of seeking far-reaching states in common motor control is much lower than that in a 'maze'. The uniform distribution policy may have trouble finding the far-reaching maze end but can drive a cheetah robot to exhibit different behaviors. Thirdly, a single exploration agent may explore very deeply, but it is hard to explore widely like a random policy. For example, a uniform distribution policy can manipulate a robotic arm to extend in all directions due to its randomness, but a single exploration agent may deeply explore only one direction. Here, a wider exploration is more important because more diverse data can help the encoder better understand motion changes. In conclusion, decoupled random collection is not good at exploring far-reaching states but still can sample a qualified dataset for visual pre-training. In addition, due to the changing exploration agent, active pre-training methods train their encoders on a changing dataset, which is unstable. By contrast, our approach decouples data sampling from encoder learning, achieving data collection before pre-training, which enables a steady learning process. Therefore, the proposed decoupled random collection is better in cross-domain pre-training due to (i) qualified pre-training data and (ii) a more stable learning process.

In summary, employing decoupled random collection brings the following advantages compared with recent advanced active pre-training methods: (i) CRPTpro has an unparalleled efficiency advantage in the pre-training stage. It employs an off-the-shelf data collection policy and doesn't need to spend time on extra RL for exploration agents. (ii) CRPTpro achieves better pre-training performance, especially when pre-training efficiency is required. Recent advanced active pre-training suffers from the severe chicken-and-egg problem mentioned above, while CRPTpro enables robust and steady learning on a qualified cross-domain pre-training dataset.

\subsection{Efficient Prototypical Learning}
\begin{figure}[t]
    \centering
    \includegraphics[scale=0.6]{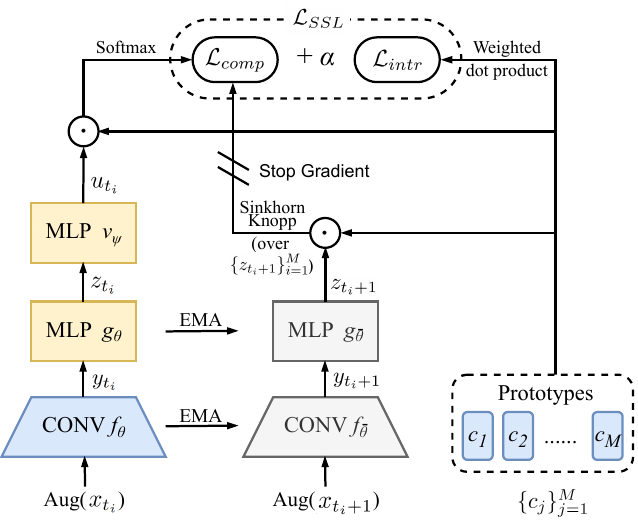}
    \caption{The proposed efficient prototypical learning in CRPTpro. It learns a visual encoder $f_\theta$ and some basic vectors called prototypes $\{c_j\}_{j=1}^M$ for downstream RL. It contains a comparative loss $\mathcal{L}_{comp}$ to compare observations projected onto prototypes (serving as cluster centers) with their clustering assignment targets, and an intrinsic loss $\mathcal{L}_{intr}$ to facilitate the coverage and diffusion of prototypes. }
    \label{ssl}
    %\vspace{-5mm}
\end{figure}

CRPTpro learns a cross-domain visual encoder and several basic vectors called prototypes over the cross-domain pre-training dataset produced by decoupled random collection. As illustrated in Fig. \ref{pipeline}, CRPTpro selects data buffers from different domains sequentially and cyclically to achieve the pre-training of one generic encoder. Concretely, in each step of the training, one data buffer is chosen to update the encoder only once, while another one will be chosen in the next step. This prevents the encoder from favoring the latest domain. 
Over the chosen buffer, CRPTpro employs efficient prototypical learning, a novel self-supervised algorithm based on prototypical representation learning. Due to the strong adaptation to small batch sizes~\cite{r24,r26}, prototypical algorithms have achieved state-of-the-art performance across different RL fields~\cite{r12,r26,r48}. As shown in Fig. \ref{ssl}, the proposed efficient prototypical learning sets several trainable prototypes as the cluster centers in the latent space, comparing observations projected onto prototypes with their clustering assignment targets. In addition, an intrinsic loss is employed to facilitate the diffusion of prototypes, along with the comparison process.

The comparative loss (in comparison) is calculated as the following. First of all, $M$ frames $\{x_{t_i}\}_{i=1}^M$ and their next frames $\{x_{t_i+1}\}_{i=1}^M$ are sampled randomly from the data buffer. The subscript represents the timing number. The current frames $\{x_{t_i}\}_{i=1}^M$ are used to predict the clustering assignment targets computed over the next frames $\{x_{t_i+1}\}_{i=1}^M$ and $M$ trainable vectors $\{c_1,...,c_M\}$ called prototypes. $x_{t_i}$ is one of the current frames. It undergoes augmentation by random image shifts, encoding by the generic convolutional encoder $f_\theta$, and projection by the Multi-Layer Perceptron (MLP) $g_\theta$ in turn to produce a vector $z_{t_i}$ in the latent space where prototypes exist. Here, another MLP $v_\psi$ is used to predict $z_{t_i}$ into $u_{t_i}$ in order to avoid collapse to trivial solutions. $v_\psi$ doesn't change the dimension of input, so both $z_{t_i}$ and $u_{t_i}$ have the same dimension as prototypes. We then take a softmax over the dot products of $u_{t_i}$ and all the prototypes $\{c_j\}_{j=1}^M$: 
\begin{equation}
\left(p_{t_i}^{(1)},...p_{t_i}^{(M)}\right) = \rm{softmax} \it \left(\frac{\hat{u}_{t_i}^T\hat{c}_1}{\tau},...,\frac{\hat{u}_{t_i}^T\hat{c}_M}{\tau}\right),
\end{equation}

\noindent where $p_{t_i}$ is the probabilities that $x_{t_i}$ maps to the prototypes for comparison, ${\tau}$ is a temperature hyper-parameter, and the hats on $u_{t_i}$ and $\{c_j\}_{j=1}^M$ denotes the $l_2$-normalization. The parameters $\theta$, $\psi$, and all prototypes are trainable and updated simultaneously when minimizing the comparative loss.

To obtain the clustering assignment target compared with $p_{t_i}$, all next frames $\{x_{t_i+1}\}_{i=1}^M$ undergo augmentation by random shifts, encoding by target encoder $f_{\Bar{\theta}}$, and projection by target MLP $g_{\Bar{\theta}}$ sequentially to produce latent embeddings $\{z_{t_i+1}\}_{i=1}^M$ analogously. To avoid collapse to trivial solutions, parameters $\Bar{\theta}$ of these target networks are updated using the Exponential Moving Average (EMA)~\cite{r32} of $\theta$:
\begin{equation}
    \Bar{\theta} \xleftarrow{} (1-\eta)\Bar{\theta} + \eta\theta. 
\end{equation}

Now we can apply the Sinkhorn-Knopp algorithm~\cite{r23} on $l_2$-normalized embeddings $\{\hat{z}_{t_i+1}\}_{i=1}^M$ and prototypes $\{\hat{c}_j\}_{j=1}^M$. Concretely, the algorithm begins with the square matrix $C$, whose elements are computed by the dot product over each embedding and prototype:

\begin{equation}
    C_{ij} = \hat{z}_{t_i+1}\hat{c}_j.
\end{equation}

Then it employs iterative doubly-normalization on the matrix $C$ to obtain target matrix $T$, constraining every column and row to have the same sum with as little change of original $C$ as possible. The row normalization and column normalization are used in doubly-normalization. The row normalization is formulated as the following: %$\rm dou(\cdot)$, the function of doubly-normalization is as the following:

\begin{equation}
    {\rm row}(C) = \frac{1}{M}C\cdot  {\rm diag}  (\frac{1}{  {\rm sum}(C,1)}),
\end{equation}
where ${\rm sum}(C,1)$ denotes the row addition and $\rm diag(\cdot)$ denotes the diagonalization of a matrix. Similarly, the column normalization is defined as the following:

\begin{equation}
    {\rm col}(C) = \frac{1}{M} {\rm diag}(\frac{1}{ {\rm sum} (C,0)}) \cdot C ,
\end{equation}
where ${\rm sum}(C,0)$ denotes the column addition. The doubly-normalization, $\rm dou(\cdot)$, consists of one row normalization and one column normalization: 

\begin{equation}
    \rm dou \it (C) = \rm col \it (\rm row \it (C)).
\end{equation}
%where ${\rm sum}(C,0)$, ${\rm sum}(C,1)$, and $\rm diag(\cdot)$ respectively denote column addition, row addition, and diagonalization of a matrix. 

Three times of doubly-normalization is applied on $C$ to obtain target matrix $T$. The $i$-th row of $T$ is the clustering assignment target $q_{t_i+1}$ of the frame $x_{t_i}$. Combined with the probabilities $p_{t_i}$ computed by Eq.(1), the comparative loss over all frames is calculated as the following:
\begin{equation}
    \mathcal{L}_{comp} = -\frac{1}{M}\sum_{i=1}^Mq_{t_i+1}^T\rm log \it p_{t_i}.
\end{equation}

\begin{algorithm}[t]
\caption{Full pseudo-code of CRPTpro.}\label{alg:alg1}
\begin{algorithmic}
\STATE
\STATE \textbf{Require:} $N$ reward-free domains $D_i$, $N$ data buffers $B_i$, $i \in [1,...N]$, $K$ downstream task MDPs $\mathcal{M}_k$, $k \in [1,...K]$, data buffer capacity $b$, pre-training update times $I_p$, (optional) finetuning update times $I_f$, and number of downstream RL steps $I_d$
\STATE \textbf{Randomly Initialize:} generic visual encoder $f_{\theta}$, projector $g_{\theta}$, predictor $v_\psi$, target encoder $f_{\Bar{\theta}}$, target projector $g_{\Bar{\theta}}$ and $M$ prototypes $\{c_j\}_{j=1}^M$.
\STATE {\textsc{\#\#\#decoupled random collection}}
\STATE \hspace{0cm}\textbf{for} $index = 1,...,b $ \textbf{do}
\STATE \hspace{0.5cm}\textbf{for} $ D_i \in[D_1,...D_N]$ \textbf{do}
\STATE \hspace{0.5cm}Take random action for $D_i$ by uniform distribution.
\STATE \hspace{0.5cm}Interact with $D_i$ to get observation $x_{index}^i$.
\STATE \hspace{0.5cm}Save $x_{index}^i$ into data buffer $B_i$.
\STATE \hspace{0.5cm}\textbf{end for}
\STATE \hspace{0cm}\textbf{end for}
\STATE {\textsc{\#\#\#efficient prototypical learning}}
\STATE \hspace{0cm}\textbf{for} $index = 1,...,I_p $ \textbf{do}
\STATE \hspace{0.5cm}Choose data buffer $B_o$, where $o=index\%N$.
\STATE \hspace{0.5cm}Randomly sample $M$ frames and next frames from $B_o$.
\STATE \hspace{0.5cm}Compute the self-supervised loss $\mathcal{L}_{SSL}$ in Eq.(10).
\STATE \hspace{0.5cm}Update $f_{\theta}$, $g_{\theta}$, $v_\psi$ and $\{c_j\}_{j=1}^M$ by backpropagation.
\STATE \hspace{0.5cm}Update $f_{\Bar{\theta}}$ and $g_{\Bar{\theta}}$ by EMA shown in Eq.(2).
\STATE \hspace{0cm}\textbf{end for}
\STATE {\textsc{\#\#\#(optional) single-domain finetuning}}
\STATE \hspace{0cm}Choose data buffer $B_f$ corresponding to target domain. If the domain is unseen, employ decoupled random collection to produce the data buffer.
\STATE \hspace{0cm}\textbf{for} $index = 1,...,I_f $ \textbf{do}
\STATE \hspace{0.5cm}Randomly sample $M$ frames and next frames from $B_f$.
\STATE \hspace{0.5cm}Compute the self-supervised loss $\mathcal{L}_{SSL}$ in Eq.(10).
\STATE \hspace{0.5cm}Update $f_{\theta}$, $g_{\theta}$, $v_\psi$ and $\{c_j\}_{j=1}^M$ by backpropagation.
\STATE \hspace{0.5cm}Update $f_{\Bar{\theta}}$ and $g_{\Bar{\theta}}$ by EMA shown in Eq.(2).
\STATE \hspace{0cm}\textbf{end for}
\STATE {\textsc{\#\#\#downstream RL in multiple domains}}
\STATE \hspace{0cm}\textbf{for} $\mathcal{M}_k \in[\mathcal{M}_1,...\mathcal{M}_K]$ \textbf{do}
\STATE \hspace{0.5cm}freeze the pre-trained $f_{\theta}$ and $\{c_j\}_{j=1}^M$.
\STATE \hspace{0.5cm}Use $f_{\theta}$ to encode the original observation space.
\STATE \hspace{0.5cm}Use $\{c_j\}_{j=1}^M$ for reward augmentation by Eq.(11).
\STATE \hspace{0.5cm}Perform $I_d$ steps RL by RAD-SAC over novel MDP.
\STATE \hspace{0cm}\textbf{end for}
\end{algorithmic}
\label{alg11}
\end{algorithm}

%According to previous work \cite{r12,r26}
When accomplishing clustering assignment tasks (optimizing the comparative loss), the prototypes serve as cluster centers for the selected samples. They are randomly initialized at the beginning and then gradually expand their coverage to the visited states in the latent space, which is done through the traction of samples \cite{r12}. These prototypes together shape the latent representation space and determine the learning effect.
Intuitively, facilitating the diffusion and coverage of prototypes can (i) accelerate the training process and (ii) make prototypes wider cluster centers, leading to a wider and more distinguishable latent space, i.e., a stronger encoder. Motivated by this, an intrinsic loss is employed along with the comparative loss, aiming to accelerate the diffusion of the prototypes, which is done by increasing the difference between prototypes. Specifically, the difference between two $l_2$-normalized prototypes (unit vectors) can be well measured by their cosine similarity:

\begin{equation}
    {\rm dif} (\hat{c}_j, \hat{c}_k) \propto -{\rm sim} (\hat{c}_j, \hat{c}_k)= -\hat{c}_j^T\hat{c}_k, 
\end{equation}
where the hats over $c_j$ and $c_k$ mean $l_2$-normalization. Our intrinsic loss is based on this cosine similarity:
\begin{equation}
    \mathcal{L}_{intr} = \sum_{j=1}^M\sum_{k=1, k\neq j}^M \frac{\rm det \it (\hat{c}_j^T) \cdot  \hat{c}_k}{\rm det \it (\hat{c}_j^T\hat{c}_k)+w},
\end{equation}
where $\rm det(\cdot)$ denotes the detach operation that prevents the gradient from backpropagation. $w$ is a constant set larger than 1, which makes the denominator %(defined as the sum of $\hat{c}_j^T\hat{c}_k$ and $w$) 
non-negative and smaller for remoter prototypes. In practice, in the case of the dot product as the loss, a remote prototype is updated slower than a near prototype after $l_2$-normalization. The smaller non-negative denominator actually increases the gradient of the remoter prototypes, thus further accelerating the diffusion of prototypes.

With the comparative loss $\mathcal{L}_{comp}$ computed by Eq.(7) and the intrinsic loss $\mathcal{L}_{intr}$ computed by Eq.(9), the overall self-supervised loss $\mathcal{L}_{SSL}$ of efficient prototypical learning is:
\begin{equation}
\mathcal{L}_{SSL} =  \mathcal{L}_{comp} + \alpha\mathcal{L}_{intr},
\end{equation}
where $\alpha$ is a weight coefficient scaling the intrinsic reward. Significantly, CRPTpro updates the parameters $\theta$, $\psi$ and prototypes $\{c_j\}_{j=1}^M$ during the pre-training. The cross-domain encoder $f_\theta$ and prototypes $\{c_j\}_{j=1}^M$ are then used to conduct efficient downstream RL in different domains.

\subsection{Downstream RL in Multiple Domains}
After pre-training, the encoder can be frozen and used to perform efficient downstream policy learning on challenging visual-control tasks from different domains either seen or unseen, with the help of frozen prototypes. Specifically, the encoder maps the state space $\mathcal{X}$ into the embedding space $\mathcal{Y}$, converting the old MDP $(\mathcal{X},\mathcal{A},\mathcal{P},\mathcal{R},\gamma,d_0)$ into $(\mathcal{Y},\mathcal{A},\mathcal{P},\mathcal{R},\gamma,d_0)$. Correspondingly, the transition $(x_{t_i},a_{t_i},r_{t_i},x_{t_i+1})$ is converted into $(y_{t_i},a_{t_i},r_{t_i},y_{t_i+1})$. Following~\cite{r12}, we augment the extrinsic reward $r_{t_i}$ by adding an exploration reward $\hat{r}_{t_i}$ based on prototypes to encourage exploration. $\hat{r}_{t_i}$ is a particle-based entropy estimation~\cite{r42} which is positively correlated with the exploration ability of the current strategy:

\begin{equation}
    \hat{r}_{t_i} = || z_{t_i+1} - z_{t_i+1}^{k\rm{NN}(Q)}||,
\end{equation}

\noindent where $z_{t_i+1}$ is the projection of $y_{t_i+1}$ in the latent space, as mentioned in Section III.B. $z_{t_i+1}^{k\rm{NN}(Q)}$ means the k-nearest neighbor of $z_{t_i+1}$ in the set $Q$. $Q$ is a projection set defined as $\{z_{t_l+1}\}$, where $z_{t_l+1}$ is chosen from the off-policy replay buffer by each prototype, according to their dot products. We refer the readers to~\cite{r12} for the detailed description of why and how $r_{t_i}$ works. With reward augmentation, the transition $(y_{t_i},a_{t_i},r_{t_i},y_{t_i+1})$ is converted into $(y_{t_i},a_{t_i},r_{t_i}+\beta\hat{r}_{t_i},y_{t_i+1})$, on which we employ RAD-SAC\cite{r31,r35} with the augmentation of random shift~\cite{r22} as the RL algorithm to learn a control policy.

In the cross-domain setting, we don't employ finetuning to maintain cross-domain versatility. However, we note that finetuning the cross-domain encoder on a single target domain is able to further improve its performance but reduce its cross-domain versatility, especially in the unseen domains. After finetuning, the encoder focuses more on a certain domain and reduces its versatility, which means CRPTpro-finetuning can be regarded as a single-domain pre-training method. We provide the full pseudo-code of CRPTpro in Algorithm \ref{alg11}.

\section{Experiments}

In this section, we verify the superiority of our CRPTpro across (i) cross-domain downstream RL performance in Section IV.B, (ii) visual generalization in Section IV.C, (iii) single-domain finetuning in Section IV.D, (iv) pre-training efficiency in Section IV.E, and (v) numerical ablation study in Section IV.F. In addition, we conduct extensive analytical experiments on both efficient prototypical learning (Section IV.G) and decoupled random collection (Section IV.H) to further verify our motivation.

\subsection{Setup}

\paragraph{Environment Details} We evaluate CRPTpro on DMControl~\cite{r13}, a representative benchmark containing many and different types of continuous visual-control tasks. It is proven and widely considered to be the most challenging motor-control benchmark for unsupervised RL~\cite{urlb,cic-nips}. According to the recent influential works~\cite{r6,r9,r12,r22}, we select 8 popular and challenging tasks defined in 8 different domains: \textit{Cheetah-Run}, \textit{Walker-Run}, \textit{Quadruped-Run}, \textit{Cartpole-Swingup sparse}, \textit{Pendulum-Swingup}, \textit{Manipulation-Reach duplo}, \textit{Hopper-Hop} and \textit{Finger-Spin}. These tasks cover different fields including balance control problems, multi-joint robot locomotion, robotic arm manipulation and so on.

\begin{table}[t]
\centering

\caption{8 challenging and diverse visual-control domains are divided into 4 different groups. The experiments of cross-domain pre-training are conducted in groups. For example, experiments on Group-A denote pre-training encoders over domains of \textit{Cheetah}, \textit{Walker}, and \textit{Quadruped}. In addition, this table intuitively shows that the selected 4 groups cover 8 domains as evenly as possible.}

\begin{tabular}{|c|cccc|}
\hline
\textit{Domain}       & Group-A              & Group-B              & Group-C              & Group-D               \\ \hline
\textit{Cheetah}      & \checkmark                    &                      &                      &                       \\
\textit{Walker}       & \checkmark                    &                      &                      & \checkmark                     \\
\textit{Quadruped}    & \checkmark                    &                      &                      &                       \\
\textit{Cartpole}     &                      & \checkmark                    &                      &                       \\
\textit{Pendulum}     & \multicolumn{1}{l}{} & \checkmark                    & \multicolumn{1}{l}{} & \checkmark                     \\
\textit{Manipulation} & \multicolumn{1}{l}{} & \checkmark                    & \checkmark                    & \multicolumn{1}{l|}{} \\
\textit{Hopper}       & \multicolumn{1}{l}{} & \multicolumn{1}{l}{} & \checkmark                    & \checkmark                     \\
\textit{Finger}       & \multicolumn{1}{l}{} & \multicolumn{1}{l}{} & \checkmark                    & \multicolumn{1}{l|}{} \\ \hline
\end{tabular}
%\vspace{-3mm}

\label{domain-groups}
\end{table}

To the best of our knowledge, we are the first to primarily concentrate on cross-domain pre-training on the challenging DMControl benchmark, whereas previous related works~\cite{r6,r9} only conduct a small number of exploratory experiments. Therefore, we divide the selected 8 domains into groups for cross-domain experiments by ourselves. They are further divided into 4 domain groups for cross-domain pre-training, as shown in Table \ref{domain-groups}. Under this division, the selected 4 groups cover the 8 domains as evenly as possible. Moreover, different groups also have different characteristics (for example, Group-A contains only multi-joint robot domains, while Group-C contains 3 different types of domains).

Following prior works, visual observations are represented as 84$\times$84$\times$3 pixel rendering and 3 consecutive previous observations are stacked to form the 84$\times$84$\times$9 state as input. Action repeat is set to 2 across all tasks. The episode length is set to 1000 for all tasks except \textit{Manipulation-Reach duplo}, with 250 episode length. %Concrete details of the chosen domains are available in Appendix A.

\begin{table*}[t]
\centering
\renewcommand\arraystretch{1.2}
\caption{Performance comparison on cross-domain pre-training. For all cross-domain methods, encoders are pre-trained on 3 domains contained in each group and then frozen to conduct downstream RL on 3 different downstream tasks defined in 3 seen domains. DrQ serves as the end-to-end expert method to demonstrate the score levels of different tasks for mean expert-normalized score calculation. CRPTpro significantly outperforms all cross-domain pre-training baselines, achieving considerable results (outperforming DrQ by 95.6\% ) across all downstream tasks.}
%\vspace{-1mm}
\setlength{\tabcolsep}{2mm}{
\begin{tabular}{|c|c|ccccc|c|}
\hline
\multirow{2}{*}{Group}   & \multirow{2}{*}{Task}             & \multicolumn{5}{c|}{Cross-domain pre-training methods}                          & End-to-end expert \\
                         &                                   & CRPTpro(ours)   & ATC \cite{r6}    & APT(C) \cite{r9}         & Proto-RL(C) \cite{r12}    & BeCL \cite{becl}           & DrQ  \cite{r22}             \\ \hline
\multirow{3}{*}{Group-A} & \textit{Cheetah-Run}              & \textbf{611±44} & 300±29  & 284±8           & 421±82          & 287±222         & 807±75            \\
                         & \textit{Walker-Run}               & \textbf{513±33} & 271±16  & 312±37          & 431±83          & 130±79          & 485±181           \\
                         & \textit{Quadruped-Run}            & 242±78          & 145±73  & 201±95          & 191±80          & \textbf{343±45} & 139±64            \\ \hline
\multirow{3}{*}{Group-B} & \textit{Cartpole-Swingup sparse}  & \textbf{722±51} & 0±0     & 20±23           & 693±43          & 529±397         & 315±243           \\
                         & \textit{Pendulum-Swingup}         & \textbf{865±26} & 524±341 & 182±98          & \textbf{868±35} & 21±9            & 635±218           \\
                         & \textit{Manipulation-Reach duplo} & \textbf{163±22} & 7±6     & 5±4             & 47±31           & 80±20           & 27±11             \\ \hline
\multirow{3}{*}{Group-C} & \textit{Manipulation-Reach duplo} & \textbf{146±31} & 8±11    & 10±10           & 37±31           & 93±24           & 27±11             \\
                         & \textit{Hopper-Hop}               & \textbf{206±4}  & 2±1     & 4±7             & 160±30          & 3±1             & 283±32            \\
                         & \textit{Finger-Spin}              & 873±157         & 883±125 & \textbf{954±23} & 858±182         & 336±425         & 938±103           \\ \hline
\multirow{3}{*}{Group-D} & \textit{Walker-Run}               & \textbf{509±60} & 200±16  & 215±5           & 433±73          & 151±96          & 485±181           \\
                         & \textit{Pendulum-Swingup}         & \textbf{875±25} & 357±249 & 86±115          & 513±294         & 22±8            & 635±218           \\
                         & \textit{Hopper-Hop}               & \textbf{210±12} & 3±3     & 22±25           & 146±32          & 2±1             & 283±32            \\ \hline
\multirow{2}{*}{-}       & Mean Score                        & \textbf{495}    & 225     & 191             & 400             & 166             & -                 \\
                         & Mean Expert-Normalized Score      & \textbf{1.956}  & 0.440   & 0.412           & 1.096           & 0.993           & -                 \\ \hline
\end{tabular}
}

\label{experiment-seen}
%\vspace{-2mm}
\end{table*}

\paragraph{Implementation of CRPTpro}
%Following the prior work ~\cite{r12}, CRPTpro employ the same neural networks architectures, where convolutional encoders $f_{\theta}$ and $f_{\Bar{\theta}}$ use stride (2,1,1,1) and filter (32,32,32,32). Projectors $g_{\theta}$ and $g_{\Bar{\theta}}$ are linear layers with 128 outputs and predictor $v_{\psi}$ is a 2-layer MLP with 1024 latent dimension and ReLU non-linearities. 
The neural networks in CRPTpro all use the same architecture from~\cite{r12}. 512 prototypes are learned by CRPTpro, each parameterized as a 128-dimensional vector.
During pre-training, the encoder is updated a total of 50k times across 3 domains. %The intrinsic loss hyper-parameters are $\alpha=5e-3$ and $w=1.5$. 
In downstream RL, 500k steps are allowed. All hyper-parameters of RAD-SAC\cite{r31,r35} are the same as~\cite{r12} except the RL reply buffer size changed from 100k to 40k. %The intrinsic reward hyper-parameters is $\beta = 0.2$ and $k$ of $k$-NN is 3. 
In both pre-training and downstream RL, Adam~\cite{r43} is chosen as optimizer with learning rate of 1e-4 and mini-batch size of 512. Results of CRPTpro are over at least 6 different evaluations (60 episodes). Table \ref{hyper} provides the settings of hyper-parameters.

\paragraph{Baselines}
4 cross-domain pre-training baselines: APT(C)~\cite{r9}, Proto-RL(C)~\cite{r12}, ATC~\cite{r6} and BeCL~\cite{becl}, 2 single-domain pre-training baselines: APT(S)~\cite{r9} and Proto-RL(S)~\cite{r12}, and 1 end-to-end image-based RL baseline: DrQ~\cite{r22} are selected through the experiment section. They are all the most recent and advanced methods, where Proto-RL(C) and Proto-RL(S) are respectively state-of-the-art cross-domain pre-training and single-domain pre-training. In particular, Proto-RL(S) is one of state-of-the-art image-based RL methods on DMControl. DrQ is a popular end-to-end method which can be employed as an expert method to show the score level of different tasks. 
%, which can be regarded as a performance upper bound for cross-domain pre-training. 
%APT(S), Proto-RL and DrQ are collectively called \textbf{‘non-cross-domain'} baselines. Their encoders are dedicated to a single domain or a single task, so they are not compared with cross-domain methods directly but serve as performance references. 

%APT pre-trains a task-agnostic agent using an entropy-driven intrinsic reward along with learning visual representations by SimCLR~\cite{r3}. 
APT~\cite{r9} learns a representation through contrastive learning by actively searching for novel states in reward-free environments. It designs a novel task-agnostic reward based on particle-based entropy maximization and trains an exploration agent over the reward to sample pre-training data. Over the sampled data, APT employs SimCLR~\cite{r3} to achieve self-supervised encoder pre-training. It is proposed for both cross-domain pre-training and single-domain pre-training, marked as APT(C) and APT(S). Proto-RL~\cite{r12} uses prototypes\cite{r24} to enhance both task-agnostic exploration and representation learning. It employs particle-based entropy maximization to train an exploration agent for data collection like APT and uses prototypes to select candidate particles better. The prototypes and visual encoder are pre-trained simultaneously by prototypical representation learning in the pre-training stage. It is the best single-domain pre-training method and a state-of-the-art visual RL algorithm on DMControl. It is marked as Proto-RL(S) in single-domain pre-training while Proto-RL(C) in cross-domain pre-training. %we port it to the cross-domain setting, marking it as 
ATC~\cite{r6} proposed an unsupervised task tailored to reinforcement learning. It requires a model to associate observations from nearby time steps within the same trajectory. Note that original ATC uses the task-specific expert dataset in pre-training and doesn't provide task-agnostic data collection method. Therefore, (i) we employ our decoupled random collection for ATC to compare its unsupervised task with our efficient prototypical learning and (ii) ATC is not included in the pre-training efficiency comparison in Section IV.E. BeCL~\cite{becl} is a recent state-of-the-art unsupervised skill discovery method. It employs contrastive learning~\cite{r3} to maximize a novel mutual information objective between observations. Unsupervised skill discovery aims to learn task-agnostic exploration, which makes it suitable for unsupervised pre-training. DrQ~\cite{r22} is a powerful and popular end-to-end DRL algorithm augmenting Q-function based on SAC~\cite{r35}. 

For cross-domain pre-training baselines, the encoder is pre-trained 50k update times (200k task-agnostic steps RL for active pre-training methods) in 3 domains and performs 500k steps downstream RL on each task like CRPTpro. For non-cross-domain baselines, we follow the settings of\cite{r12}: 500k steps (i.e., 125k update times) task-agnostic active pre-training in one domain and 500k steps downstream RL on each task, for APT(S) and Proto-RL(S); 1M steps task-specific end-to-end RL on each task for DrQ.% More details of baselines are provided in Appendix A. 

%This demonstrates that our CRPTpro is the state-of-the-art cross-domain encoder pre-training method. Its huge advantages on the mean over other cross-domain baselines also verifies this conclusion. 

%Compared with non-cross-domain references, CRPTpro achieves competitive RL performance with much less pre-training consumption. It outperforms APT(S) across all tasks and DrQ on 8/12 tasks, achieving 8.216 and 1.956 Performance Ratio respectively. Perhaps the most exciting result is that CRPTpro achieves a 0.992 Performance Ratio to Proto-RL, demonstrating that CRPTpro reaches the same level as the best image-based RL method on DMControl, plus the added bonus of an encoder that is generic across multiple domains. 
%This suggests on visual-control tasks, the cross-domain diversity can effectively offset the exploration shortage of the random exploration policy. More analysis and visualization are available in Appendix B.
%This suggests that even in the single-domain pre-training where the chicken-and-egg problem is not severe, our decoupled random collection also supports competitive pre-training, where the cross-domain diversity is the main contributor. More analysis and visualization are available in Appendix B.

\begin{table*}[t]
\centering
\renewcommand\arraystretch{1.2}
\caption{Performance comparison of cross-domain methods generalizing to unseen domains and unseen colors without finetuning. Pre-trained encoders are frozen and used to conduct 500k steps RL on unseen downstream tasks defined in unseen domains (or domains with unseen colors). CRPTpro overall outperforms all cross-domain baselines.} 
%\vspace{-1mm}
\setlength{\tabcolsep}{3mm}{
\begin{tabular}{|cccccc|}
\hline
\multicolumn{1}{|c|}{Task}                                                                    & CRPTpro(ours)                           & ATC \cite{r6}                            & APT(C) \cite{r9}                                  & Proto-RL(C) \cite{r12}                    & BeCL \cite{becl}                                    \\ \hline
\multicolumn{6}{|l|}{(i) Encoders are pre-trained on Group-A and tested on 5 tasks from 5 unseen domains.}                                                                                                                                                                                    \\ \hline
\multicolumn{1}{|c|}{\textit{Pendulum-Swingup}}                                               & \textbf{671±273}                        & 214±159                        & 502±346                                 & 118±75                         & 23±8                                    \\
\multicolumn{1}{|c|}{\textit{Finger-Spin}}                                                    & 809±232                                 & \textbf{879±133}               & 748±47                                  & 850±194                        & 3±1                                     \\
\multicolumn{1}{|c|}{\textit{Hopper-Hop}}                                                     & \textbf{187±19}                         & 23±27                          & 36±31                                   & 138±47                         & 2±1                                     \\
\multicolumn{1}{|c|}{\textit{Manipulation-Reach duplo}}                                       & \textbf{62±18}                          & 10±13                          & 18±19                                   & 20±7                           & \textbf{64±37}                          \\
\multicolumn{1}{|c|}{\textit{Cartpole-Swingup sparse}}                                        & \textbf{93±17}                          & 20±12                          & 15±12                                   & 39±24                          & 0±0                                     \\ \hline
\multicolumn{6}{|l|}{{\color[HTML]{000000} (ii) Encoders are pre-trained on Group-C and tested on 5 tasks from 5 unseen domains.}}                                                                                                                                                            \\ \hline
\multicolumn{1}{|c|}{{\color[HTML]{000000} \textit{Cheetah-Run}}}                             & {\color[HTML]{000000} \textbf{448±12}}  & {\color[HTML]{000000} 241±10}  & {\color[HTML]{000000} 283±31}           & {\color[HTML]{000000} 374±10}  & {\color[HTML]{000000} 350±223}          \\
\multicolumn{1}{|c|}{{\color[HTML]{000000} \textit{Walker-Run}}}                              & {\color[HTML]{000000} \textbf{570±26}}  & {\color[HTML]{000000} 268±8}   & {\color[HTML]{000000} 231±33}           & {\color[HTML]{000000} 497±75}  & {\color[HTML]{000000} 139±67}           \\
\multicolumn{1}{|c|}{{\color[HTML]{000000} \textit{Quadruped-Run}}}                           & {\color[HTML]{000000} 270±32}           & {\color[HTML]{000000} 132±66}  & {\color[HTML]{000000} 114±63}           & {\color[HTML]{000000} 234±40}  & {\color[HTML]{000000} \textbf{322±123}} \\
\multicolumn{1}{|c|}{{\color[HTML]{000000} \textit{Pendulum-Swingup}}}                        & {\color[HTML]{000000} \textbf{872±27}}  & {\color[HTML]{000000} 135±102} & {\color[HTML]{000000} 778±106}          & {\color[HTML]{000000} 524±370} & {\color[HTML]{000000} 23±10}            \\
\multicolumn{1}{|c|}{{\color[HTML]{000000} \textit{Cartpole-Swingup sparse}}}                 & {\color[HTML]{000000} \textbf{682±98}}  & {\color[HTML]{000000} 0±1}     & {\color[HTML]{000000} 9±10}             & {\color[HTML]{000000} 94±79}   & {\color[HTML]{000000} 0±0}              \\ \hline
\multicolumn{6}{|l|}{{\color[HTML]{000000} (iii) Encoders are pre-trained on Group-C and tested on 3 tasks with unseen background color.}}                                                                                                                                                    \\ \hline
\multicolumn{1}{|c|}{{\color[HTML]{000000} \textit{Manipulation-Reach duplo (unseen color)}}} & {\color[HTML]{000000} \textbf{163±16}}  & {\color[HTML]{000000} 34±20}   & {\color[HTML]{000000} 15±9}             & {\color[HTML]{000000} 55±52}   & {\color[HTML]{000000} 5±6}              \\
\multicolumn{1}{|c|}{{\color[HTML]{000000} \textit{Pendulum-Swingup (unseen color)}}}         & {\color[HTML]{000000} \textbf{567±317}} & {\color[HTML]{000000} 136±101} & {\color[HTML]{000000} \textbf{556±104}} & {\color[HTML]{000000} 126±37}  & {\color[HTML]{000000} 21±9}             \\
\multicolumn{1}{|c|}{{\color[HTML]{000000} \textit{Cheetah-Run (unseen color)}}}              & {\color[HTML]{000000} \textbf{485±26}}  & {\color[HTML]{000000} 184±18}  & {\color[HTML]{000000} 174±21}           & {\color[HTML]{000000} 337±36}  & {\color[HTML]{000000} 1±1}              \\ \hline
\end{tabular}}

\label{experiment-unseen}
%\vspace{-5mm}
\end{table*}

%\vspace{-2mm}
\subsection{Cross-domain Pre-training}
We compare CRPTpro with 4 cross-domain pre-training baselines over 4 groups defined in Table \ref{domain-groups}. CRPTpro hyper-parameter settings are shown in Table \ref{hyper}. DrQ is set as an end-to-end expert to show the score level of different downstream tasks and calculate an normalized score. Results are summarized in Table \ref{experiment-seen}. CRPTpro significantly improves APT(C), ATC and BeCL on 11/12 tasks. Compared with Proto-RL(C), CRPTpro achieves better downstream policy learning on 11/12 tasks and similar performance on the rest task. Numerically, we calculate the mean score and the mean expert-normalized score over all 12 downstream tasks of 4 groups, where our CRPTpro improves upon the next best cross-domain pre-training baseline by 23.8\% and 78.5\% respectively. Perhaps the most exciting result is that CRPTpro achieves a 1.956 mean expert-normalized score, demonstrating the huge advantages of cross-domain visual pre-training for image-based RL, which is ignored by previous works. 

In summary, our CRPTpro significantly beats all baselines, becoming a novel state-of-the-art cross-domain pre-training method. This is attributed to the following reasons: First, our decoupled random collection avoids the severe chicken-and-egg problem between exploration agent training and visual encoder pre-training (detailed description in Section III.A), generating a qualified and diverse cross-domain pre-training dataset and enabling stable pre-training process. Second, our efficient prototypical learning helps the encoder learn more effective image embeddings. 

%\vspace{-3mm}
\subsection{Generalization in Unseen Domains}
In addition to the seen domains, the encoder pre-trained by CRPTpro can generalize well to unseen domains without finetuning. We conduct three sets of representative experiments to show the generalization comparison of all cross-domain methods. (i) We pre-train encoders on Group-A which contains only robot domains. Then we freeze the pre-trained encoders to directly perform 500k steps downstream RL on the rest tasks defined in five unseen and different types of domains: manipulation (\textit{Manipulation}), robot (\textit{Hopper}), and balance control (\textit{Pendulum}, \textit{Cartpole}, and \textit{Finger}). (ii) We pre-train encoders on Group-C which contains three different types of domains: manipulation (\textit{Manipulation}), robot (\textit{Hopper}) and balance control (\textit{Finger}), and then freeze the encoder to perform 500k steps RL on the other five unseen tasks. (iii) For encoders pre-trained on Group-C, we further change the background color (by exchanging the RGB value in reverse order) of three different types of control tasks from both the seen domain (\textit{Manipulation}) and unseen domains (\textit{Cheetah} and \textit{Pendulum}), testing the generalization to unseen background colors. CRPTpro hyper-parameters are provided in Table \ref{hyper}.

The results are shown in Table \ref{experiment-unseen}. CRPTpro overall exceeds all cross-domain baselines across all three sets of experiments. In some unseen tasks (e.g., \textit{Pendulum-Swingup} in (i) and \textit{Walker-Run} in (ii)), CRPTpro even enables better policy learning than the end-to-end expert DrQ. Note that CRPTpro has not trained its encoder in these unseen domains, while DrQ trains its encoder for 1M steps on each task. This indicates that the encoder of CRPTpro can well capture the movement changes that are generic across domains. Moreover, CRPTpro is the only method that doesn't exhibit performance degradation on \textit{Cheetah-Run} when facing unseen color, which means it can effectively filter our useless information irrelevant to movements. We mainly attribute these movement understanding advantages to the proposed decoupled random collection because we note that CRPTpro on Group-C (containing three different types of domains) performs much better than CRPTpro on Group-A (containing only robot domains). For example, CRPTpro on Group-C can effectively solve all unseen tasks, including \textit{Cartpole-Swingup sparse}, while CRPTpro on Group-A can not. Even on \textit{Walker-Run} which is included in Group-A but not in Group-C, CRPTpro on Group-C performs better than CRPTpro on Group-A (result in Table \ref{experiment-seen}). This phenomenon shows that data diversity is the key to pre-training performance, which is also verified in the ablation study (Section IV.F). The more diverse dataset sampled by decoupled random collection enables CRPTpro to better understand generic movement changes than all baselines. %Although the performance is not satisfactory enough in domain \textit{Cartpole}, it could also be easily improved through few-shot finetuning, which we details in next section.

%Take experiment (i) as an example, CRPTpro conducts effective downstream policy learning on \textit{Pendulum-Swingup}, \textit{Finger-Spin}, \textit{Hopper-Hop} and \textit{Manipulation-Reach duplo}, even achieving better performance than end-to-end expert DrQ on two of them (DrQ is shown in Table \ref{experiment-seen}. 

\begin{figure*}[t]
    \centering
    \includegraphics[scale=0.13]{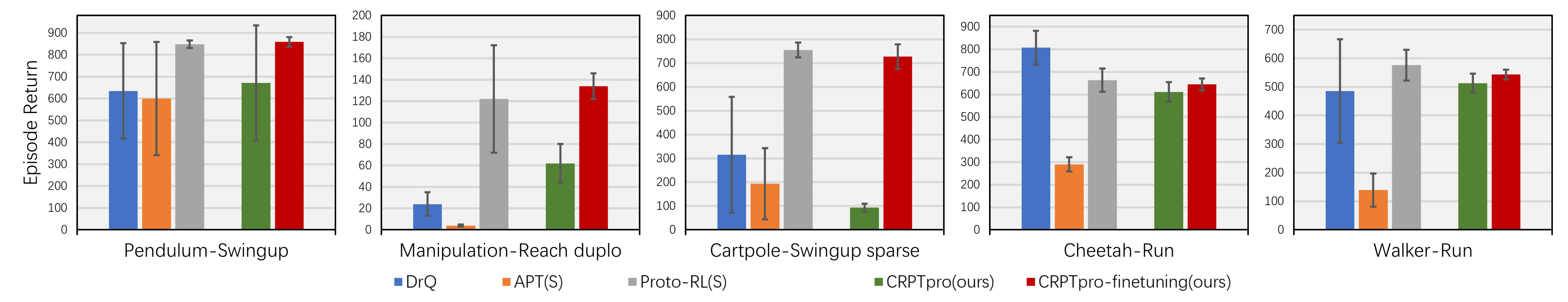}
    %\vspace{-2mm}
    \caption{ Performance comparison between CRPTpro-finetuning and non-cross-domain baselines. The encoder is pre-trained by CRPTpro on Group-A and then finetuned on 3 unseen domains and 2 seen domains respectively. Finetuning is helpful on both seen domains and unseen domains. CRPTpro-finetuning could be regarded as a single-domain pre-training method, achieving competitive downstream performance with state-of-the-art single-domain pre-training Proto-RL(S) which is also one of state-of-the-art image-based RL methods. }
    \label{finetuning-all}
    %\vspace{-2mm}
\end{figure*}

%\vspace{-3mm}
\subsection{Finetuning}
As mentioned in Section III.C, the cross-domain encoder obtained by CRPTpro can be finetuned on a target domain for better downstream policy learning, especially in unseen domains. To highlight this ability, we finetune the encoder pre-trained on Group-A by CRPTpro in five different domains (three unseen domains: \textit{Pendulum}, \textit{Manipulation}, and \textit{Cartpole} \& two seen domains: \textit{Cheetah} and \textit{Walker}) respectively with default hyper-parameters in Table \ref{hyper}. Then, we freeze the finetuned encoder to conduct 500k steps RL on the corresponding downstream task. CRPTpro-finetuning could be regarded as a single-domain pre-training method and we compare it with three non-cross-domain baselines.% (including two single-domain pre-training methods and an end-to-end RL method).
%To highlight this ability, we pre-train the encoder on Group-A and then finetune it on 4 domains, of which 3 are seen and 1 is unseen.

The results are shown in Fig. \ref{finetuning-all}, demonstrating that CRPTpro-finetuning is an effective single-domain pre-training method. It is comparable with Proto-RL(S), the best single-domain pre-training method and also one of state-of-the-art image-based RL methods. In addition, CRPTpro-finetuning uses only 58.5k update times on encoder (50k pre-training update times on Group-A, 2k finetuning update times in \textit{Pendulum}, 2k finetuning update times in \textit{Manipulation}, 3k finetuning update times in \textit{Cartpole}, 1k finetuning update times in \textit{Cheetah}, and 0.5k finetuning update times in \textit{Walker}) to surpass most non-cross-domain baselines. As a comparison, the total pre-training update times for Proto-RL(S) and APT(S) in 5 domains is 625k. In each unseen domain, CRPTpro only updates its encoder at most 3k times while the baselines update their encoders at least 125k times. This indicates that the cross-domain prior knowledge learned with our novel self-supervised algorithm can greatly promote representation learning even in unseen domains. 

\begin{table}[t]
\centering
\caption{Comparison between CRPTpro and five pre-training algorithms on pre-training efficiency. Outcomes come from experiments conducted on the NVIDIA Tesla P100. CRPTpro is the most efficient pre-training algorithm.}
%\vspace{-1mm}
\renewcommand\arraystretch{1.1}
\setlength{\tabcolsep}{0.9mm}{
\begin{tabular}{|cccccccl|}
\hline
                                                                                                    &                                & \begin{tabular}[c]{@{}c@{}}CRPTpro\\ (ours)\end{tabular} & \begin{tabular}[c]{@{}c@{}}Proto-RL\\ (C)\end{tabular} & BeCL         & \begin{tabular}[c]{@{}c@{}}APT\\ (C)\end{tabular} & \begin{tabular}[c]{@{}c@{}}Proto-RL\\ (S)\end{tabular} & \begin{tabular}[c]{@{}l@{}}APT\\ (S)\end{tabular} \\ \hline
\multicolumn{1}{|c|}{\multirow{3}{*}{\begin{tabular}[c]{@{}c@{}}Wall-\\ clock\\ time\end{tabular}}} & \multicolumn{1}{c|}{1 domain}  & \textbf{4.8h}                                            & 8.8h                                                   & 9.4h         & 12.2h                                             & 18.8h                                                  & 20.6h                                             \\
\multicolumn{1}{|c|}{}                                                                              & \multicolumn{1}{c|}{2 domains} & \textbf{4.8h}                                            & 8.8h                                                   & 9.4h         & 12.2h                                             & 37.6h                                                  & 41.2h                                             \\
\multicolumn{1}{|c|}{}                                                                              & \multicolumn{1}{c|}{3 domains} & \textbf{4.8h}                                            & 8.8h                                                   & 9.4h         & 12.2h                                             & 56.4h                                                  & 61.8h                                             \\ \hline
\multicolumn{1}{|c|}{\multirow{3}{*}{\begin{tabular}[c]{@{}c@{}}Update\\ times\end{tabular}}}       & \multicolumn{1}{c|}{1 domain}  & \textbf{50k}                                             & \textbf{50k}                                           & \textbf{50k} & \textbf{50k}                                      & 125k                                                   & 125k                                              \\
\multicolumn{1}{|c|}{}                                                                              & \multicolumn{1}{c|}{2 domains} & \textbf{50k}                                             & \textbf{50k}                                           & \textbf{50k} & \textbf{50k}                                      & 250k                                                   & 250k                                              \\
\multicolumn{1}{|c|}{}                                                                              & \multicolumn{1}{c|}{3 domains} & \textbf{50k}                                             & \textbf{50k}                                           & \textbf{50k} & \textbf{50k}                                      & 375k                                                   & 375k                                              \\ \hline
\end{tabular}}

\label{pre-training efficiency}
%\vspace{-3mm}
\end{table}

\subsection{Pre-training Efficiency}
%Due to the usage of stochastic dataset, CSPTpro is much more efficient in the pre-training stage compared with previous active pre-training methods. 
%In this section, we demonstrate the pre-training efficiency comparison between 3 prototypical pre-training methods when facing different numbers of domains. They are Proto-RL(S), the current state-of-the-art single-domain pre-training method; Proto-RL(C), the current state-of-the-art cross-domain pre-training method; and our CRPTpro which outperforms current state-of-the-art cross-domain pre-training method. The outcomes are based on experiments conducted on the NVIDIA Tesla P100. 
In this section, we demonstrate the pre-training efficiency comparison between different pre-training methods when facing different numbers of domains. Three cross-domain pre-training methods, including the current state-of-the-art cross-domain pre-training method, Proto-RL(C), and two single-domain pre-training methods, including the current state-of-the-art single-domain pre-training method, Proto-RL(S), are employed for comparison with our CRPTpro. ATC is not in comparison because it doesn't provide a task-agnostic pre-training approach in their paper (see baseline details in Section IV.A). The outcomes are based on experiments conducted on the NVIDIA Tesla P100.

\begin{figure}[t]
    \centering
    %\vspace{-3mm}
    \includegraphics[width=0.47\textwidth]{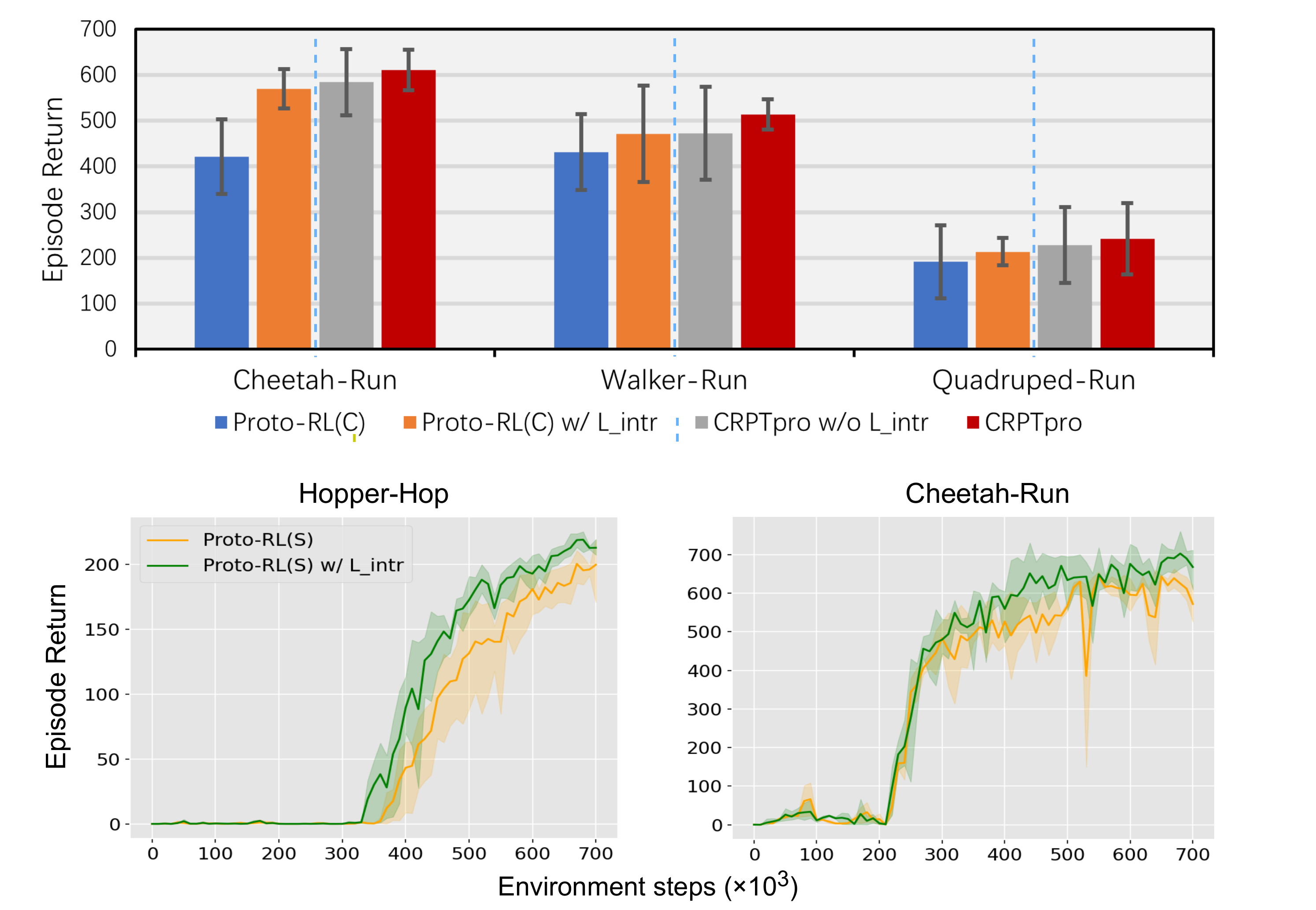}
    %\vspace{-3mm}
    \caption{  \textbf{Up:} Ablating $\mathcal{L}_{intr}$ from CRPTpro and adding $\mathcal{L}_{intr}$ into Proto-RL(C). This figure also serves as the ablation study of both the decoupled random collection and efficient prototypical learning in CRPTpro. \textbf{Down:} Adding $\mathcal{L}_{intr}$ into Proto-RL(S). Efficient prototypical learning is effective when employed in 3 different pre-training settings.}
    \label{lossintr}
    %\vspace{-2mm}
\end{figure}

The results are shown in Table \ref{pre-training efficiency}. CRPTpro becomes a novel state-of-the-art cross-domain pre-training method with greatly improved pre-training efficiency. Concretely, it spends only 54\% wall-clock pre-training time to outperform the next best Proto-RL(C) by 78.5\% on the mean expert-normalized score. The main contributor is our decoupled random collection, which employs an off-the-shelf exploration policy to avoid the severe training burden of extra RL. Compared with the state-of-the-art single-domain pre-training method (Proto-RL(S)), all the cross-domain pre-training methods exhibit huge efficiency improvements which multiply with the number of domains. This is because cross-domain pre-training methods can pre-train a generic encoder across multiple domains, while Proto-RL(S) cannot. However, the downstream policy performance of all cross-domain baselines is much worse than that of Proto-RL(S), which makes the efficiency comparison between them meaningless. In contrast, CRPTpro can achieve competitive downstream policy learning compared with Proto-RL(S) after few-shot finetuning (see Section IV.D), with much less pre-training consumption. 

In downstream RL, CRPTpro augments the extrinsic reward with a k-NN-based unsupervised reward. According to our tests on the NVIDIA Tesla P100, the consumption of this reward is tiny, about 1‰ of the downstream RL consumption. This is due to (i) the small size of the k-NN buffer and (ii) the low dimension of the k-NN particles, which we detail in Section III.C. In addition, the current state-of-the-art Proto-RL(C) and Proto-RL(S) also utilize this reward. Therefore, it doesn't affect the efficiency advantage of our approach.

\begin{figure}[t]
\centering
\subfloat{
  \includegraphics[width=0.235\textwidth]{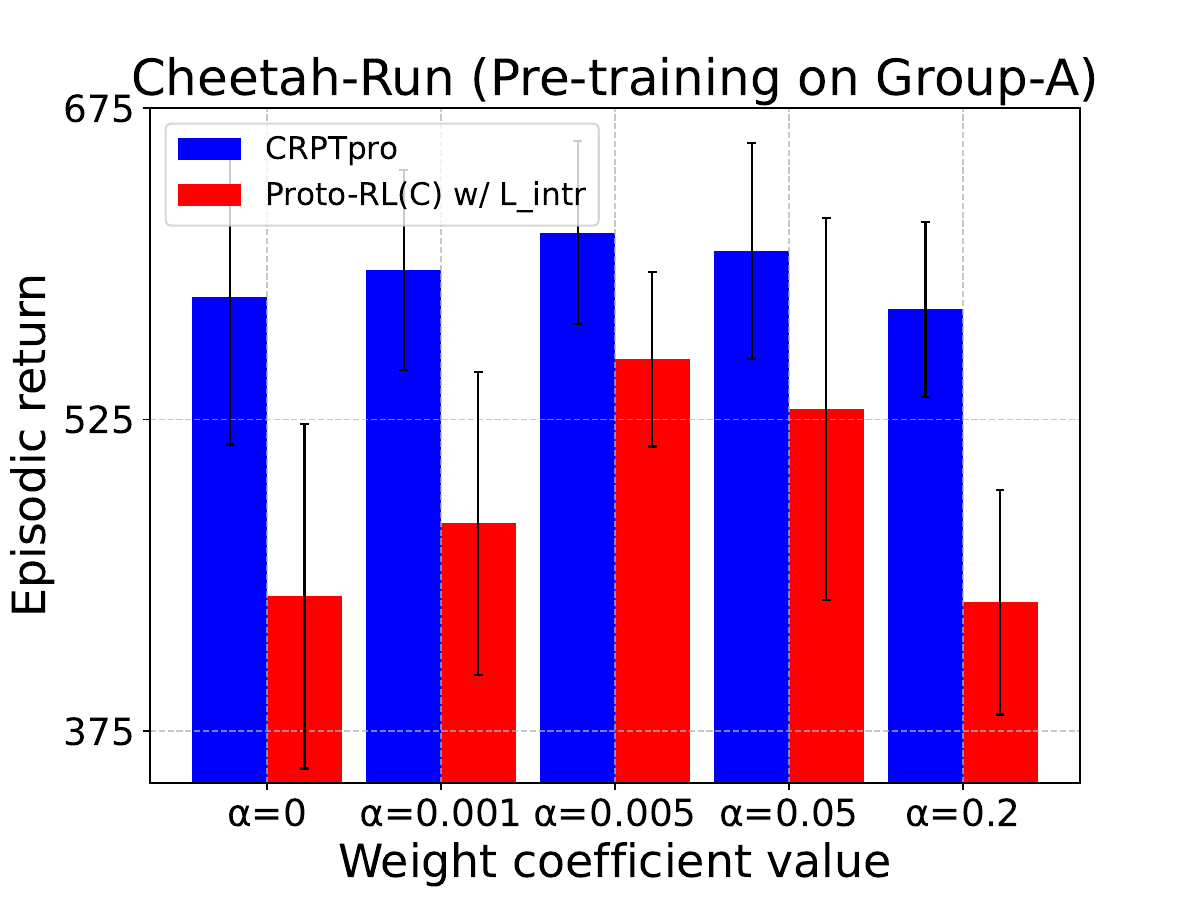}
  \label{subfig:first}
}
\hspace{-0.03\textwidth} % 调整这个值来改变间距
\subfloat{
  \includegraphics[width=0.245\textwidth]{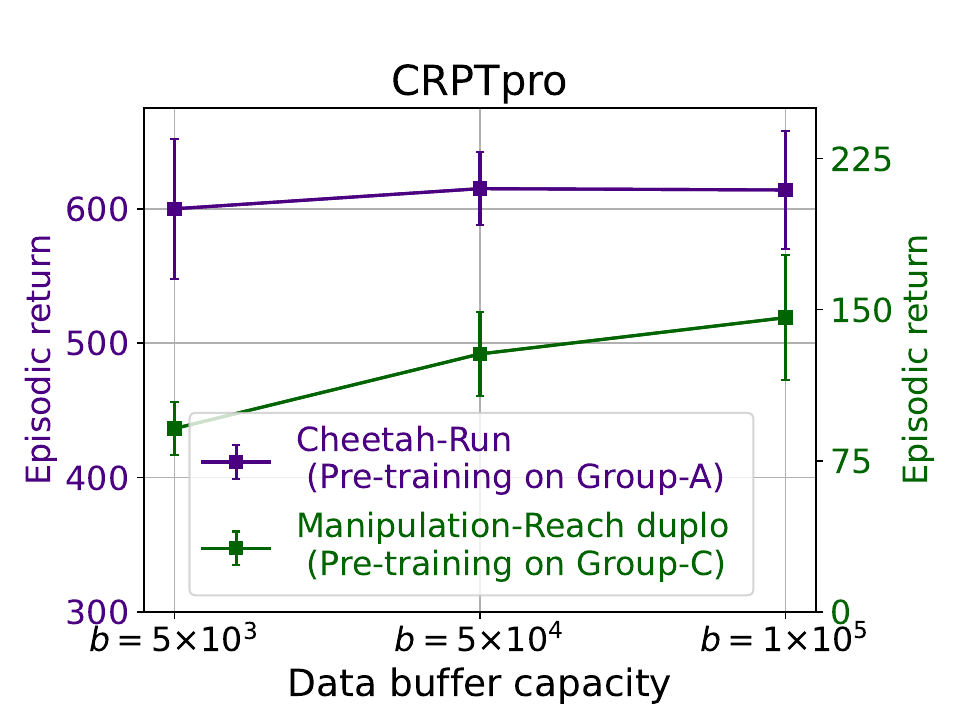}
  \label{subfig:second}
}

\caption{ The hyper-parameter sensitivity analysis of $\mathcal{L}_{intr}$ weight coefficient $\alpha$ in two cross-domain pre-training methods (\textbf{left}) and data buffer capacity $b$ in decoupled random collection of CRPTpro on two tasks (\textbf{right}).}
\label{e12}
%\vspace{-2mm}
\end{figure}

\subsection{Numerical Ablation Study} 
%CRPTpro employs the proposed efficient prototypical learning, which introduces a novel $\mathcal{L}_{intr}$ to facilitate the diffusion of prototypes and improve pre-training. In this section, we demonstrate its effectiveness and rationality through three different sets of experiments
In this section, we conduct extensive experiments to verify the necessity and robustness of our proposed decoupled random collection and efficient prototypical learning.

First, we test the intrinsic loss $\mathcal{L}_{intr}$ in three different prototypical pre-training settings. The ablation study (both the decoupled random collection and efficient prototypical learning) is also done here. % where prototypical learning is employed. 
%\paragraph{Cross-domain static pre-training}
(i) We ablate the $\mathcal{L}_{intr}$ from CRPTpro with default hyper-parameter settings to observe the difference, shown in Fig. \ref{lossintr} up. 
%All other hyper-parameters of CSPTpro(w/o $\mathcal{L}_{intr}$) remains the same as CSPTpro.
%The result in fig.4(a) demonstrates that $\mathcal{L}_{intr}$ is necessary for CSPTpro. 
%\paragraph{Cross-domain active pre-training}
(ii) We add $\mathcal{L}_{intr}$ into Proto-RL(C) with default hyper-parameter settings to show its effectiveness
%The intrinsic loss hyper-parameters are $\alpha=5e-3$ and $w=1.5$ with other hyper-parameters unchanged. 
in Fig. \ref{lossintr} up. %The improvement demonstrates $\mathcal{L}_{intr}$ is also helpful for prototypical learning in cross-domain active pre-training settings.
%\paragraph{Single-domain active pre-training}
(iii) We add $\mathcal{L}_{intr}$ into Proto-RL(S) of 200k steps task-agnostic pre-training and 500k steps downstream RL. The settings are all shown in Table \ref{hyper} except $\alpha$ set to $1e-3$ for \textit{Cheetah-Run} and $5e-4$ for \textit{Hopper-Hop}. The curves are shown in Fig. \ref{lossintr} down.
%In Proto-RL, $\mathcal{L}_{intr}$ is able to facilitate prototypical learning, achieving better downstream policy learning when pre-training steps are not sufficient.
The numerical results verify the effectiveness of $\mathcal{L}_{intr}$ in efficient prototypical learning across all prototypical pre-training settings. Since the Proto-RL(C) w/ $\mathcal{L}_{intr}$ is actually CRPTpro w/o decoupled random collection, Fig. \ref{lossintr} up can also serve as the ablation study, verifying the effectiveness of both the decoupled random collection and efficient prototypical learning in CRPTpro.

Then, we further conduct the sensitivity analysis on two key hyper-parameters: $\mathcal{L}_{intr}$ weight coefficient $\alpha$ (used in efficient prototypical learning) and random data buffer capacity $b$ (used in decoupled random collection). In each experiment, all the other hyper-parameters are default. The results shown in Fig. \ref{e12} left demonstrate that $\mathcal{L}_{intr}$ is helpful and robust to weight coefficient $\alpha$ when it is not set too large. This is consistent with our intuition because $\mathcal{L}_{intr}$ is set to assist the original clustering assignment task, where the comparative loss $\mathcal{L}_{comp}$ is the target. $\mathcal{L}_{comp}$ is computed by prototypes (serving as clustering centers) and encoders, while the prototypes and encoders are also shaped by $\mathcal{L}_{comp}$. If $\alpha$ is too large, the prototypes and $\mathcal{L}_{comp}$ will be overly affected, leading to unstable encoder learning and performance drops. For data buffer capacity $b$ (Fig. \ref{e12} right), even though CRPTpro can still pre-train an effective encoder (better than all cross-domain baselines) with a small dataset, a larger buffer is a better choice if there is enough memory. In summary, both the effectiveness and robustness of proposed methods are further verified.

\begin{table}[b]
\caption{ Employing efficient prototypical learning facilitates the diffusion of prototypes and makes them cover wider after pre-training.}
\centering
\renewcommand\arraystretch{1.1}
\setlength{\tabcolsep}{0.75mm}{
\begin{tabular}{|c|cc|cc|cc|}
\hline
($\times10^{-4}$)    & CRPTpro          & \begin{tabular}[c]{@{}c@{}}CRPTpro\\ w/o $\mathcal{L}_{intr}$\end{tabular} & \begin{tabular}[c]{@{}c@{}}Proto-RL\\ (C)\\ w/ $\mathcal{L}_{intr}$\end{tabular} & \begin{tabular}[c]{@{}c@{}}Proto-RL\\ (C)\end{tabular} & \begin{tabular}[c]{@{}c@{}}Proto-RL\\ (S)\\ w/ $\mathcal{L}_{intr}$\end{tabular} & \begin{tabular}[c]{@{}c@{}}Proto-RL\\ (S)\end{tabular} \\ \hline
KNE $\downarrow$     & \textbf{-26.545} & -10.871                                                                    & \textbf{-38.602}                                                                 & -7.701                                                 & \textbf{-80.131}                                                                 & -44.036                                                \\
ANE $\downarrow$     & \textbf{7.996}   & 23.778                                                                     & \textbf{21.635}                                                                  & 51.280                                                 & \textbf{20.587}                                                                  & 48.774                                                 \\ \hline
$\mathcal{L}_{intr}$ & \textbf{-31.080} & -20.505                                                                    & \textbf{-50.294}                                                                 & -26.860                                                & \textbf{-140.213}                                                                & -80.839                                                \\ \hline
\end{tabular}}
\label{ane}
\end{table}

\begin{figure}[t]
    \centering
    \includegraphics[width=0.47\textwidth]{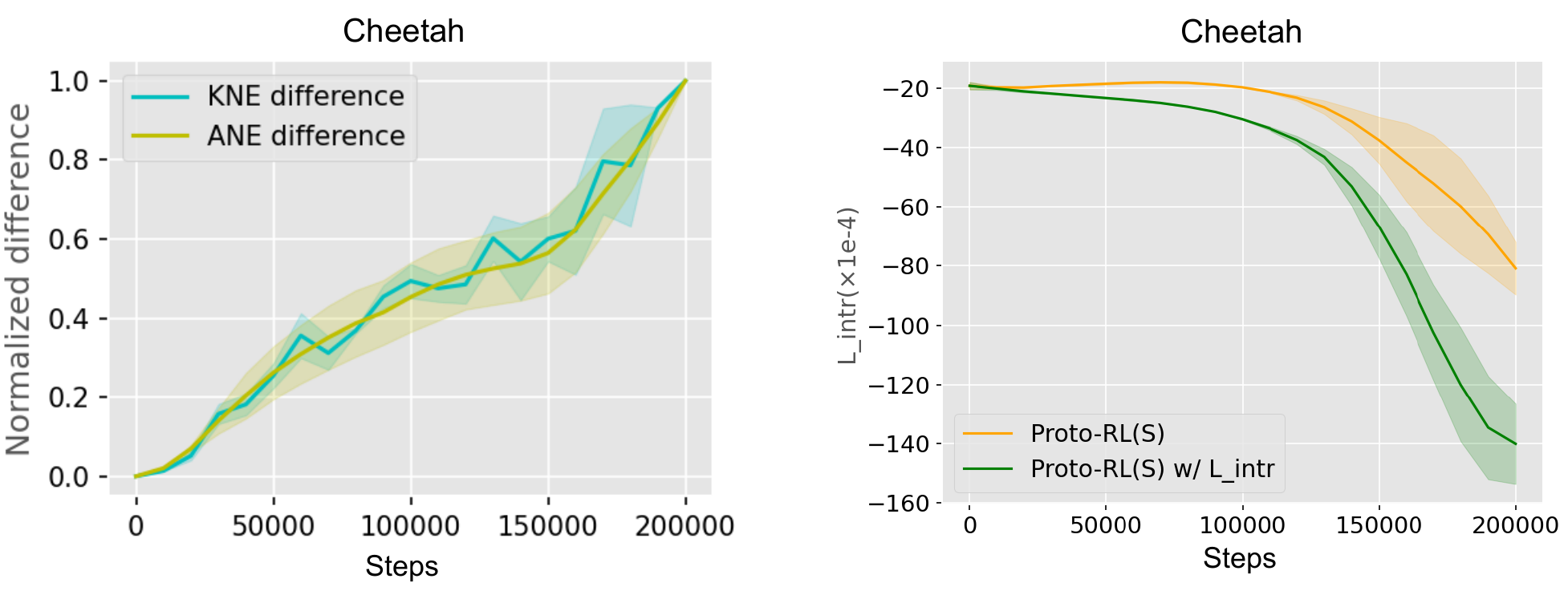}
    %\vspace{-2mm}
    \caption{\textbf{Left:} The $\mathcal{L}_{intr}$ continues to function throughout the training process of efficient prototypical learning, making prototypes cover faster and wider. \textbf{Right:} Optimizing $\mathcal{L}_{intr}$ is an instinctual process in prototypical algorithms, which motivates us to optimize $\mathcal{L}_{intr}$ in a targeted manner.  }
    \label{lossintr_assist}
    %\vspace{-2mm}
\end{figure}

\subsection{Analysis of Efficient Prototypical Learning}

In addition to numerical ablations, we further conduct several analysis experiments on efficient prototypical learning.

First, we show that the coverage and diffusion of prototypes are facilitated when employing efficient prototypical learning. Two metrics are used to evaluate prototypes' coverage: All-Neighbor Estimation (ANE) proportional to the cosine similarity between all prototype pairs; K-Neighbor Estimation (KNE) proportional to k-nearest neighbor's cosine similarity of all prototypes. We test the pre-trained models' prototypes in 3 different prototypical pre-training settings. Results in Table \ref{ane} show that $\mathcal{L}_{intr}$ makes prototypes finally cover wider in the latent space. In addition, we show the curve of metric difference between Proto-RL(S) and Proto-RL(S) w/ $\mathcal{L}_{intr}$ during pre-training in Fig. \ref{lossintr_assist} left. This indicates that the $\mathcal{L}_{intr}$ continues to function throughout the training process of efficient prototypical learning, demonstrating that $\mathcal{L}_{intr}$ makes prototypes cover faster in the self-supervised pre-training. In summary, our efficient prototypical learning indeed improves the diffusion and coverage of prototypes, leading to better encoder pre-training and the final performance.

Then, we observe that optimizing $\mathcal{L}_{intr}$ is an instinctual process in vanilla prototypical algorithms. We show the $\mathcal{L}_{intr}$ curve in the single-domain active pre-training setting, as shown in Fig. \ref{lossintr_assist} right. We find that even in Proto-RL(S), where $\mathcal{L}_{intr}$ is not employed, this loss is also instinctively decreased, which means reducing $\mathcal{L}_{intr}$ is an instinctual process in vanilla prototypical representation learning. This phenomenon motivates us to optimize $\mathcal{L}_{intr}$ in a targeted manner to facilitate the training and accelerate the convergence.

\subsection{Analysis of Decoupled Random Collection}

In addition to numerical ablations, we further conduct several analysis experiments on decoupled random collection.

\begin{figure}[t]
\centering
\subfloat{
  \includegraphics[width=0.24\textwidth]{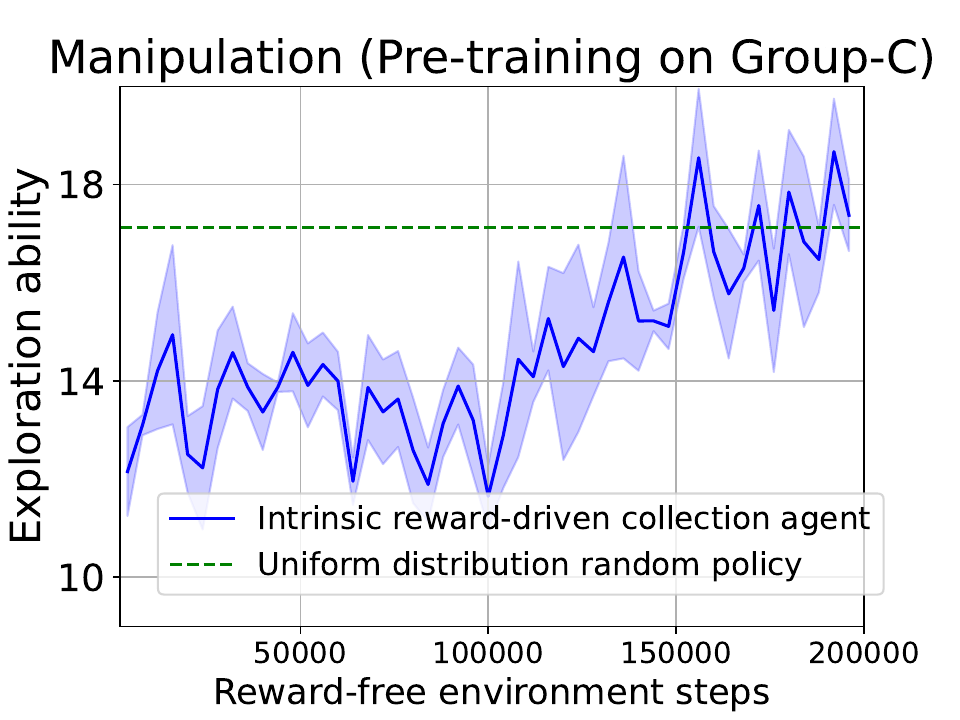}
  \label{subfig:first}
}
%\hfill % 水平填充，可以使图形相互间有一些间隔
\subfloat{
  \includegraphics[width=0.24\textwidth]{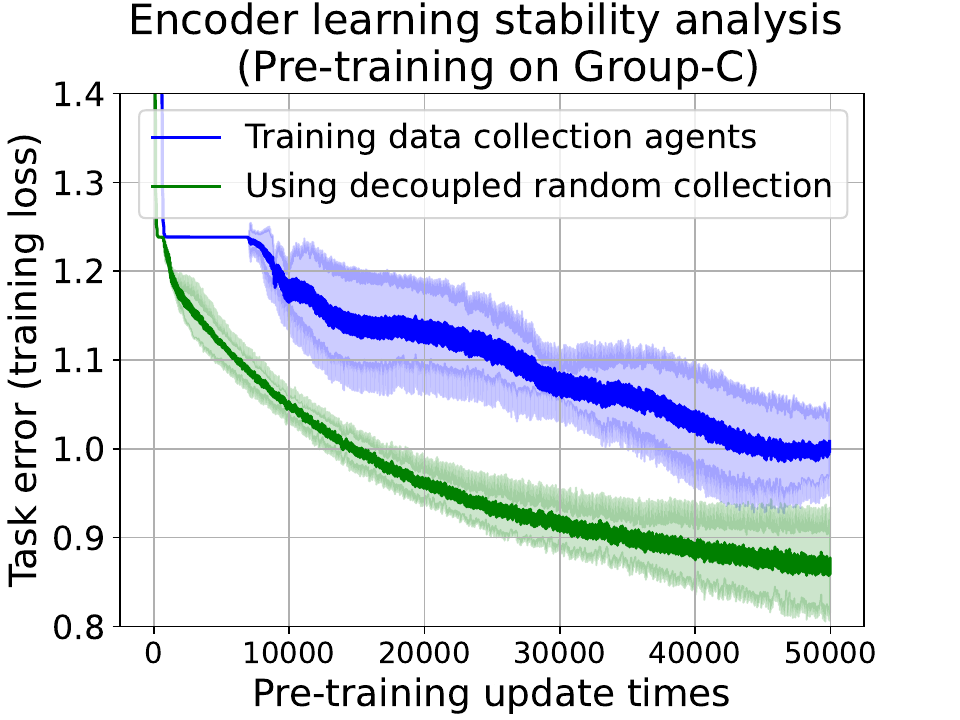}
  \label{subfig:second}
}
%\vspace{-3mm}
\caption{ \textbf{Left:} The exploration ability comparison of intrinsic reward-driven agents and uniform random policy in cross-domain settings. \textbf{Right:} The effect of using decoupled random collection on self-supervised task error curves.}
\label{e34}
%\vspace{-3mm}
\end{figure}

First, we show the exploration ability comparison between our uniform random policy (CRPTpro) and intrinsic reward-driven collection agents (Proto-RL(C)) during the whole pre-training process. The exploration ability is evaluated by the coverage of sampled data, which is estimated by a particle-based estimator (based on k-nearset-neighbor distance) introduced by \cite{r9,r12}. The results shown in Fig. \ref{e34} left demonstrate that the training of exploration policies suffers from low sample efficiency in cross-domain settings. Its exploration exceeds that of the uniform policy only towards the end of the pre-training, which is attributed to the severe chicken-and-egg conflict between multiple exploration policy training and cross-domain representation learning. By contrast, the uniform random policy can indeed provide considerable exploration and stable pre-training data all the time without training. 

Second, we analyze the encoder training stability effect of using decoupled random collection. In pre-training, encoders are updated through accomplishing self-supervised tasks. We employ decoupled random collection instead of training data collection agents in Proto-RL(C), observing how its self-supervised clustering assignment task error (i.e., the comparative training loss $\mathcal{L}_{comp}$) changes during pre-training. The error (loss) curves are shown in Fig. \ref{e34} right. The decoupled random collection enables a more efficient optimization process, i.e., a more steady encoder learning process. It avoids the aforementioned chicken-and-egg conflict plaguing active pre-training methods, achieving stable representation learning on a static and qualified dataset of considerable exploration.

\begin{figure}[t]
    \centering
    \includegraphics[width=0.4\textwidth]{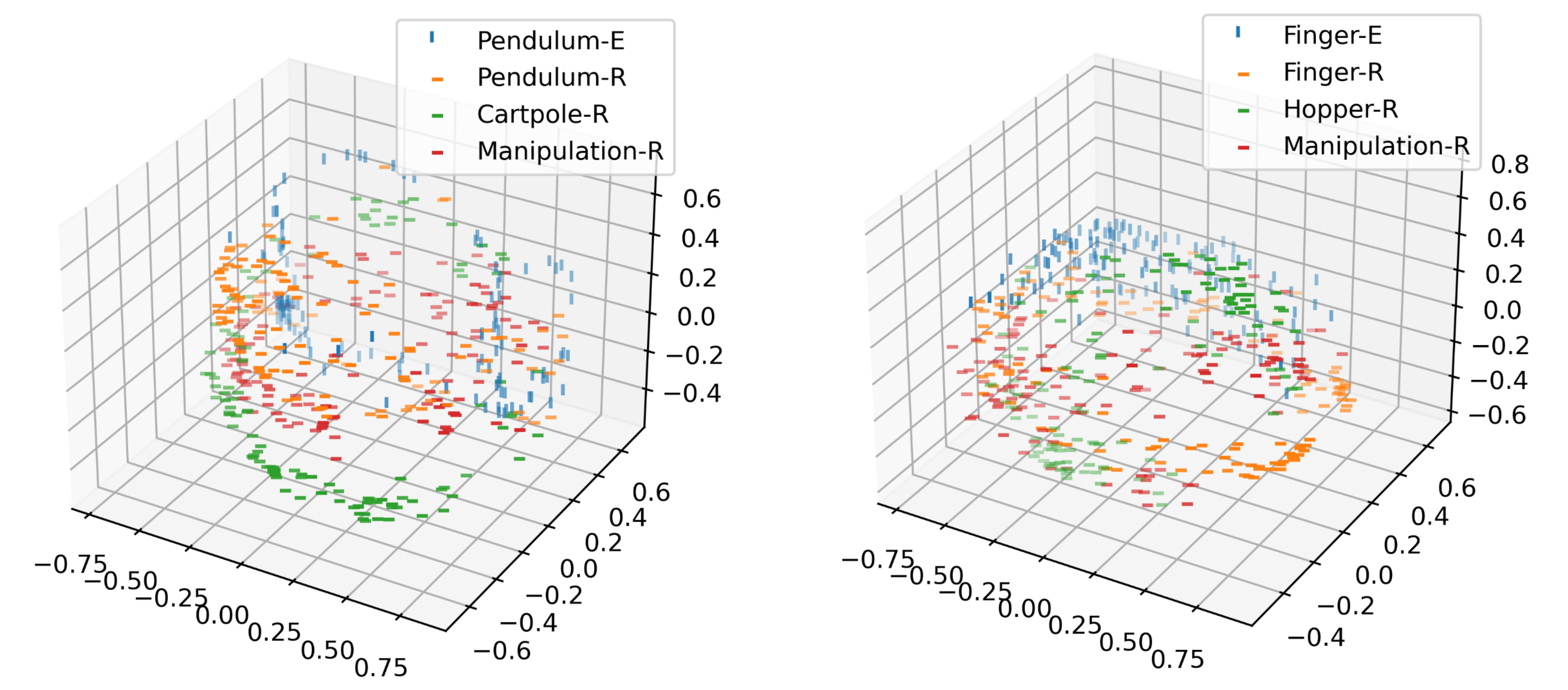}
    %\vspace{-3mm}
    \caption{ Visualization of the cross-domain pre-training dataset produced by decoupled random collection (CRPTpro) and the single-domain exploration dataset produced by Proto-RL(S). %The data are encoded by corresponding CRPTpro encoder and then reduced dimensionality to 4 by PCA (4-dimensional PCA results retain more than half of the original information here). 
    In the legend, `\textbf{E}' after domains denotes the exploration agents of Proto-RL(S) and `\textbf{R}' after domains denotes the decoupled random collection of CRPTpro. %The triad under each sub image denotes which 3 dimensions of 4-dimensional PCA results are visualized. 
    \textbf{Left:} Proto-RL(S) on domain \textit{Pendulum} and the decoupled random collection on Group-B. %The explained variance ratio of PCA results is (0.230,0.145,0.109,0.061). 
    \textbf{Right:} Proto-RL(S) on domain \textit{Finger} and the decoupled random collection on Group-C. %The explained variance ratio of PCA results is (0.268,0.137,0.099,0.066). 
    }
    \label{visual}
    %\vspace{-3mm}
\end{figure}

%To further analyze the effectiveness of our decoupled random collection and pre-trained encoder, 
Finally, we visualize the cross-domain pre-training dataset produced by decoupled random collection (CRPTpro) and the single-domain exploration dataset produced by Proto-RL(S)~\cite{r12}. Note that Proto-RL(S) provides the state-of-the-art unsupervised exploration. We don't visualize Proto-RL(C) here because its performance is much worse than CRPTpro and Proto-RL(S). For Proto-RL(S), we conduct adequate single-domain pre-training (500k steps) before collecting data for visualization. All the data are encoded by a well-trained CRPTpro encoder and reduced dimensionality to 3 by PCA for visualization. Results are shown in Fig. \ref{visual}. In a single domain, the uniform policy (orange) is not as exploratory as the exploration agents (blue), but not a lot worse. It also explores some observations that Proto-RL(S) cannot reach. Moreover, the cross-domain random dataset (orange, green and red) overall exhibits wider coverage than the single-domain exploration dataset (blue) because of the cross-domain diversity. In addition, we observe that our encoder can ideally handle unseen data sampled by Proto-RL(S) and distinguish them clearly, showing the powerful generalization. These explain why CRPTpro can pre-train a generic encoder conducting efficient downstream policy learning in multiple domains, even reaching the same level as the best active pre-training method (also a state-of-the-art image-based RL method), Proto-RL(S), after few-shot finetuning.

%To further analyze the effectiveness of our decoupled random collection and pre-trained encoder, we employ PCA to visualize the latent embeddings of two pre-training datasets: the cross-domain pre-training dataset produced by decoupled random collection in CRPTpro and the single-domain exploration dataset produced by exploration agents in Proto-RL(S)~\cite{r12}. Note that Proto-RL(S) is a state-of-the-art image-based RL method, providing the state-of-the-art unsupervised exploration. Concretely, we sample 375 frames from the cross-domain pre-training dataset (125 frames from each domain) and 125 frames from the single-domain exploration dataset, and then we use the encoder pre-trained by CRPTpro to encode all the sampled frames into latent embeddings where PCA is employed. We use the first 4 dimensions of the PCA results (Retaining more than half of the original information) and perform 3D visualizations in Group-B and Group-C.

\section{Conclusion}

In this paper, we present CRPTpro, a novel cross-domain RL pre-training framework enabling state-of-the-art downstream policy learning on sets of visual-control tasks defined in different domains. The proposed decoupled random collection successfully circumvents the chicken-and-egg conflict (between exploration policy training and representation learning), which plagues current advanced active pre-training. Due to the upper limits of random policies, we hope this work could inspire novel thinking of cross-domain RL pre-training paradigms (e.g., sample-efficient cross-domain active pre-training or novel cross-domain pre-training frameworks). Moreover, our success demonstrates that the random data contains much useful information. It may also be helpful in other fields, such as state-based skill discovery (e.g., providing more state-action information) or learning from videos (e.g., providing action supervision when extracting actions from expert videos). In addition, we hope the proposed efficient prototypical learning can encourage more research on clustering-based self-supervised algorithms in RL.

\begin{table}[h]
\caption{Hyper-parameter settings in CRPTpro.}
\centering
\renewcommand\arraystretch{1.1}
\setlength{\tabcolsep}{2mm}{
\begin{tabular}{|l|c|}
\hline
Parameter                      & Setting       \\ \hline
Convolution channels             & $(32,32,32,32)$ \\
Convolution stride             & $(2,1,1,1)$     \\
Filter size                    & $3\times3$           \\
Representation dimensionality  & $39200$         \\
Latent dimensionality          & $128$           \\
Number of prototypes $M$         & $512$           \\
Predictor hidden units         & $1024$          \\
Feature dimensionality   & $50$            \\
MLP hidden units         & $1024$          \\
%Critic feature dimensionality  & $50$            \\
%Critic MLP hidden units        & $1024$          \\
Batch size                      & $512$         \\
Optimizer                   & Adam          \\
Learning rate                  & $10^{-4}$         \\
Random data buffer capacity $b$                 & $100000$          \\
Intrinsic loss weight coefficient $\alpha$    & $5\times10^{-3}$         \\
Encoder target update frequency     & $1$         \\
Encoder target EMA momentum     & $0.05$         \\
%RL optimizer                   & Adam          \\
%RL learning rate                  & $10^{-4}$          \\
RL algorithm                   & RAD-SAC       \\
Random shift pad               & $\pm4$             \\
SAC initial temperature        & $0.1$           \\
Discount factor $\gamma$                      & $0.99$          \\
SAC replay buffer capacity                       & $40000$          \\
Actor update frequency         & $2$             \\
Actor log stddev bounds        & $[-10,2]$   \\
Critic update frequency        & $1$             \\
Critic target update frequency & $2$             \\
Critic target EMA momentum     & $0.01$          \\
Exploration reward $k$      & $3$             \\
Exploration reward coefficient $\beta$   & $0.2$           \\
Exploration reward buffer size   & $2048$          \\
Constant $w$        & $1.5$           \\
Softmax temperature ${\tau}$            & $0.1$           \\ \hline
\end{tabular}
}
\label{hyper}
%\vspace{-5mm}
\end{table}

%\section*{Acknowledgments}
%This should be a simple paragraph before the References to thank those individuals and institutions who have supported your work on this article.

%{\appendix%[Full hyper-parameters of CRPTpro]
%The hyper-parameter settings of CRPTpro are shown in Table \ref{hyper}. 

%We provide the full hyper-parameters of our proposed CRPTpro which is shown in Table \ref{hyper}.


\bibliographystyle{IEEEtran}
\bibliography{ecai}

\vfill

\end{document}